\documentclass[lettersize,journal]{IEEEtran}
\usepackage{amsmath,amsfonts}
\usepackage{algorithmic}
\usepackage{algorithm}
\usepackage{array}
\usepackage[caption=false,font=normalsize,labelfont=sf,textfont=sf]{subfig}
\usepackage{textcomp}
\usepackage{stfloats}
\usepackage{url}
\usepackage{verbatim}
\usepackage{graphicx}
\usepackage{cite}
\usepackage{color}
\usepackage{cleveref}
\crefname{figure}{Fig.}{Fig.}
\crefname{section}{Section}{Section}
\crefname{table}{TABLE}{TABLE}
\crefname{equation}{Eq.}{Eq.}

\usepackage{multirow}
\usepackage{booktabs} 
\usepackage{pifont}

\begin{document}

\title{Accurate Pose Estimation for Flight Platforms based on Divergent Multi-Aperture Imaging System}

\author{Shunkun Liang, Bin Li, Banglei Guan, Yang Shang, Xianwei Zhu, Qifeng Yu
	\thanks{Manuscript received ** *, 2024; revised ** *, 2024; accepted * *, 2024. Date of publication * *, 2024; date of current version * *, *. The Associate Editor coordinating the review process was ** *. (Corresponding authors: Banglei Guan, guanbanglei12@nudt.edu.cn; Yang Shang, shangyang1977@nudt.edu.cn)}
	\thanks{Shunkun Liang, Bin Li, Banglei Guan and Yang Shang are with the College of Aerospace Science and Engineering, National University of Defense Technology, Changsha 410000, China (e-mail: guanbanglei12@nudt.edu.cn; shangyang1977@nudt.edu.cn)
		
	Xianwei Zhu is with the Institute of Intelligent Optical Measurement and Detection, Shenzhen University, Shenzhen 518000, China (e-mail: zxw2899@szu.edu.cn)
		
	Qifeng Yu is with the College of Aerospace Science and Engineering, National University of Defense Technology, Changsha 410000, China (e-mail: yuqifeng@nudt.edu.cn).
	}
}

\markboth{\tiny{This work has been submitted to the IEEE for possible publication. Copyright may be transferred without notice, after which this version may no longer be accessible.}}%
{Shell \MakeLowercase{\textit{et al.}}: A Sample Article Using IEEEtran.cls for IEEE Journals}


\maketitle

\begin{abstract}
Vision-based pose estimation plays a crucial role in the autonomous navigation of flight platforms. However, the field of view and spatial resolution of the camera limit pose estimation accuracy. This paper designs a divergent multi-aperture imaging system (DMAIS), equivalent to a single imaging system to achieve simultaneous observation of a large field of view and high spatial resolution. The DMAIS overcomes traditional observation limitations, allowing accurate pose estimation for the flight platform. {Before conducting pose estimation, the DMAIS  must be calibrated. To this end  we propose a calibration method for DMAIS based on the 3D calibration field.} The calibration process determines the imaging parameters of the DMAIS, {which allows us to model DMAIS as a generalized camera.}
Subsequently, a new algorithm for accurately determining the pose of flight platform is introduced. We transform the absolute pose estimation problem into a nonlinear minimization problem. New optimality conditions are established for solving this problem based on Lagrange multipliers. Finally, real calibration experiments show the effectiveness and accuracy of the proposed method. Results from real flight experiments validate the system's ability to achieve centimeter-level positioning accuracy and arc-minute-level orientation accuracy.
\end{abstract}


\section{Introduction}
\label{sec:introduction}
Visual technologies have developed rapidly in recent years and have been applied across various fields\cite{Yu2024,guan2022monitoring,Liu2024TIP}. Accurate pose estimation of the flight platform in space plays an important role in various missions, {such as navigation\cite{Lu2018,Wang2021} and target localization\cite{Qian2024,Liu2024TIP}}. Especially in benchmark alignment applications, accurate pose estimation guides the flight platform to follow a predefined trajectory. While satellite and inertial navigation have traditionally been used for pose estimation in outdoor environments\cite{Du2022,Li2024}, {they each have inherent limitations}. {Satellite navigation may be unreliable due to external signal disruptions,} while inertial navigation suffers from error accumulation over time. In contrast, vision-based pose estimation technology presents several advantages. It is immune to time drift and error accumulation, operates independently of external signals, and is self-sufficient. By processing image data, this technology can accurately determine the platform's position and orientation in real time, thereby facilitating precise movement guidance\cite{He2024,Lu2018}.
\begin{figure}[tbp]
	\centering
	\includegraphics[width=0.95\linewidth]{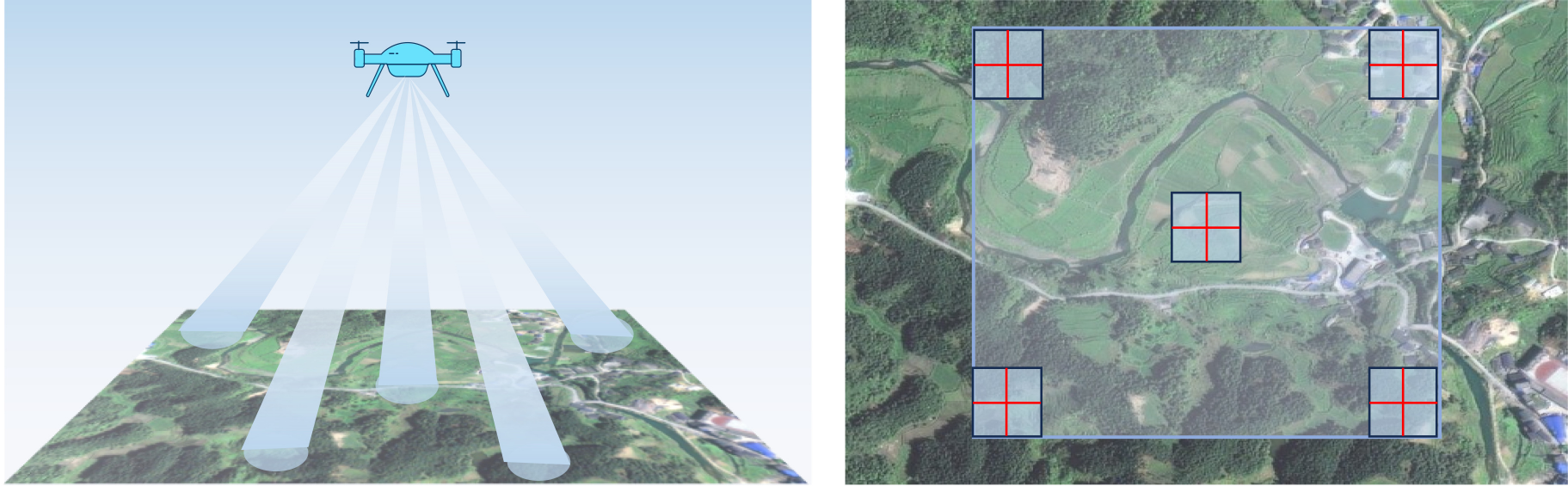}
	\caption{Multi-aperture imaging is equivalent to simultaneous large field of view and high spatial resolution imaging.}
	\label{fig:5cam}
\end{figure}

In recent years, there has been significant interest and application in vision-based pose estimation for flight platforms\cite{Patel2020,Yol2014,Goforth2019}. {Relative pose estimation focuses on the transformation between two observed images \cite{guan2023minimal,Kneip2014,Yu2023}.} In contrast, absolute pose estimation is concerned with determining the pose of the imaging system within the world coordinate\cite{Kneip2013,Upnp2014}. This paper focuses on the application of DMAIS in estimating flight platform pose in the world coordinate. A common method is to perform feature matching between the images captured by the camera on flight platform and the pre-stored image sequence with known structure, and then use the visual geometry theory to estimate the absolute pose\cite{Goforth2019,Wang2023,Patel2020, He2024}. Gofor et al.\cite{Goforth2019} introduce a method to align an image captured by a single camera with the pre-existing satellite image to estimate the pose of an unmanned air vehicle (UAV). Similarly, Patel et al. \cite{Patel2020} {obtain} reference images from Google Earth and use stereo cameras along with an inertial measurement unit (IMU) to determine the six degrees of freedom (DoF) global pose of UAV. 

Vision-based pose estimation methods have advanced significantly, yet achieving higher accuracy remains challenging. High-accuracy pose estimation requires the imaging system to have both high spatial resolution and large field of view (FoV). However, current imaging hardware cannot {simultaneously meet these dual requirements}. In the case of single camera\cite{Goforth2019} or stereo camera system\cite{Patel2020}, spatial resolution and FoV share an inverse relationship, {making it unfeasible to achieve both high spatial resolution and a wide FoV concurrently.} To address this limitation, this paper proposes a divergent multi-aperture imaging system (DMAIS), which can provide accurate pose estimation for the flight platform. Unlike conventional multi-aperture systems \cite{Qi2022,Laycock2007}, the designed DMAIS consists of five long-focal cameras, each with a dispersed observation orientation and no overlapping field of view. This unique configuration allows each camera to maintain a high spatial resolution with a narrow FoV. When combined, the divergent orientation of these cameras effectively simulates a large FoV, as shown in \cref{fig:5cam}. Consequently, the DMAIS enables visual pose estimation with both a large FoV and a high spatial resolution, equivalent to a singular imaging system achieving these feats simultaneously. 

To achieve accurate pose estimation, two fundamental issues must be addressed: (1) geometric calibration of DMAIS and (2) absolute pose estimation algorithm based on DMAIS. Geometric calibration aims to determine the intrinsic and extrinsic parameters of DMAIS. {The purpose of absolute pose estimation is to calculate the pose of the DMAIS in the world frame from the captured images}.
The specific analysis is as follows:

\textbf{(1) Geometrical calibration of the DMAIS.} 
The long-focal cameras, characterized by their narrow FoV and extended observation distances, present significant challenges when using planar target calibration methods\cite{Zhang2000,Sturm1999}. A common method is {constructing} a 3D calibration field with a predefined structure\cite{Oniga2018,Shang2013,Liu2015}. Specifically, Liu et al.\cite{Liu2015} propose a method to calibrate the camera by distributing multiple high-precision small planar targets in the FoV of the camera. Oniga et al.\cite{Oniga2018} lay markers on buildings to calibrate cameras on UAV. The absence of overlapping FoVs among cameras complicates the calibration of extrinsic parameters. {Liang et al. \cite{liang2025camera} propose arranging pattern on the focal plane of the collimator, which can be considered as points from an infinite distance.} Some studies have proposed mirror-based methods\cite{Takahashi2012,Long2015}. Such methods are impractical for the long-focus multi-camera system due to the impracticality of constructing sufficiently large flat mirrors. Infrastructure-based methods\cite{Heng2014,Choi2018,Lin2020} {require prior reconstruction of the scene structure.} However, the cost of large-scale scene reconstruction is huge. {This paper proposes an intrinsic and extrinsic calibration method based on a 3D calibration field. We use the control points from the calibration field to estimate the intrinsic parameters of each long-focus camera.} Subsequently, by treating the central camera as the reference frame, {we establish connections between the cameras using a multi-rotation strategy. Finally, we compute the extrinsic parameters of each camera in a unified reference frame.} 

\textbf{(2) Absolute pose estimation algorithm based on DMAIS.} Matching images from the calibrated DMAIS with the pre-stored image sequence having the known structure \cite{Nassar2018,Sun2023,He2024} allows us to establish correspondences between the 2D image observations and the 3D world point can be obtained. Given a set of 2D-3D correspondences, estimating the absolute pose of the camera is a fundamental problem in photogrammetry. Until now, numerous absolute pose estimators have been developed, including EPnP\cite{Lepetit2009}, DLS\cite{Hesch2011}, OPnP\cite{Zheng2013}, UPnP\cite{Kneip2014}, optDLS\cite{Nakano2015} and SQPnP\cite{Terzakis2020}. Hesch et al.\cite{Hesch2011} {formulate the absolute pose problem as a minimization problem, and calculate the camera pose by solving the first-order optimality conditions.} Based on DLS\cite{Hesch2011}, Laurent et al.\cite{Kneip2014} {propose} a Unified PnP (UPnP) approach applicable for both central and non-central cameras. Furthermore, Nakano \cite{Nakano2015} {introduces new optimality conditions with Lagrangian multipliers, and demonstrates the efficiency of Cayley parameterization for rotation.} However, optDLS\cite{Nakano2015} is limited to single-camera absolute pose estimation. This paper models the DMAIS as a generalized camera and extends optDLS\cite{Nakano2015} to the multi-camera system. Real flight experiments show that our method can achieve accurate platform pose estimation.

The paper is structured as follows: \cref{sec:DMAIS} introduces the divergent multi-aperture imaging system. \cref{sec:calib} details the geometric calibration methods used for the intrinsic and extrinsic parameters of the DMAIS. In the \cref{sec:pose}, the DMAIS is modeled as a generalized camera, and an absolute pose estimation algorithm is proposed. \cref{sec:result} reports the real-world data experiment results to evaluate the calibration and pose estimation methods. Finally, \cref{sec:conclusions} concludes the paper.

\section{Divergent Multi-Aperture Imaging System}
\label{sec:DMAIS}

Accurate visual pose estimation typically requires the imaging system to have both high spatial resolution and a large FoV. {A larger field of view (FoV) provides stronger spatial geometric constraints and pose discrimination capabilities, thereby enhancing the accuracy of pose estimation. The higher spatial resolution allows for capturing more details of the scene. This capability enhances the ability to recognize features within the scene, ultimately improving pose estimation accuracy.} Current hardware technology limits a single camera's ability to fulfill both requirements due to the inverse relationship between these two attributes. Especially, for an imaging system with a pixel resolution $(P)$, as the FoV $(W)$ increases, the value of $(W/P)$ increases and the spatial resolution's level decreases correspondingly. To address this challenge, we design a divergent multi-aperture imaging system for accurate pose estimation of the flight platforms. 

\begin{figure}[htbp]
	\centering
	\includegraphics[width=0.45\linewidth]{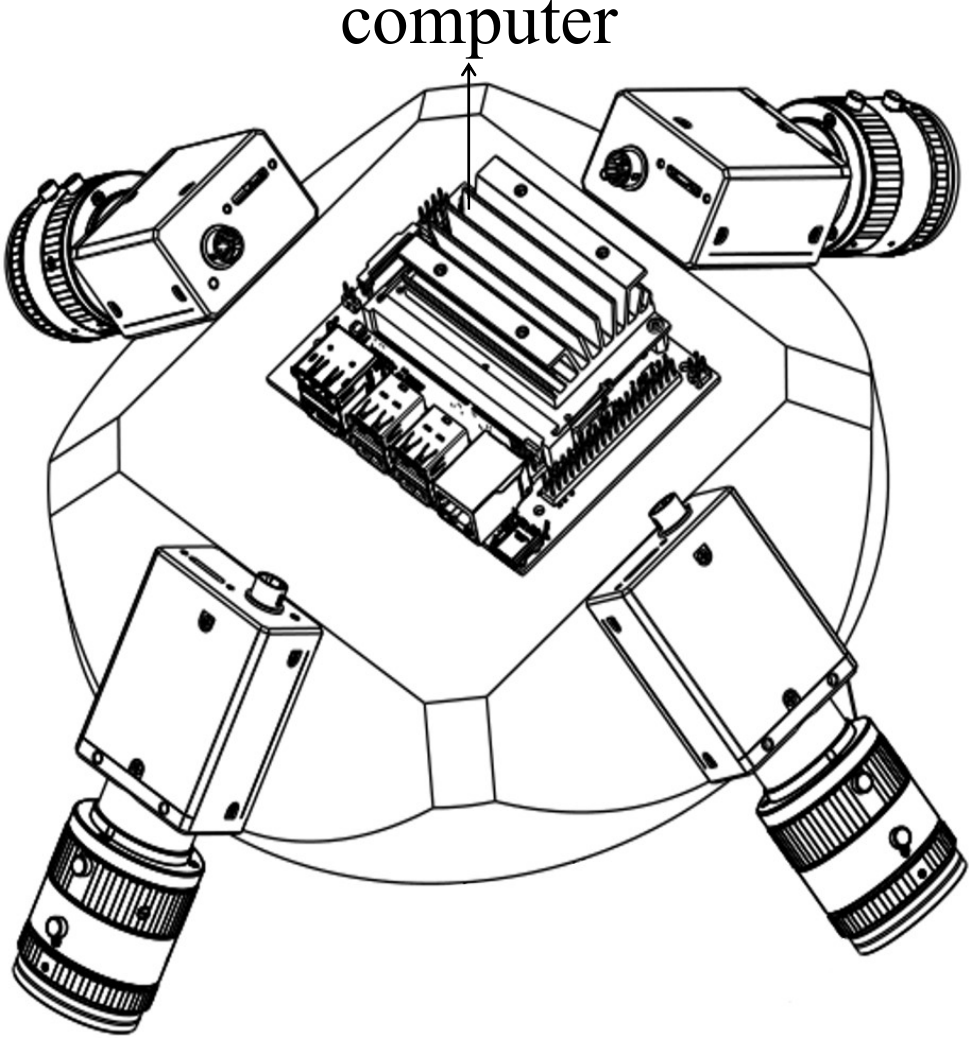}
	\includegraphics[width=0.45\linewidth]{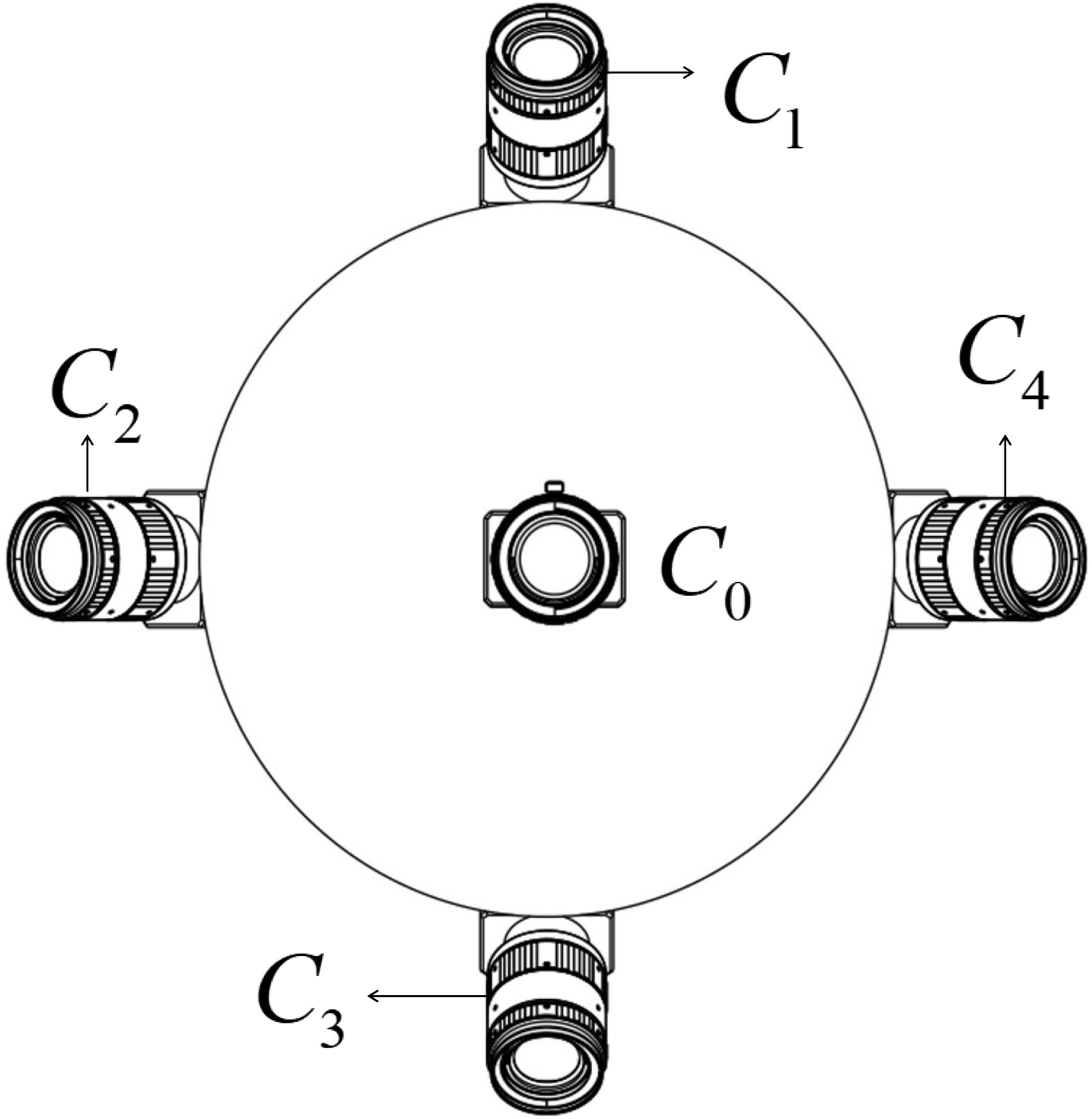}
	\caption{Structure diagram of divergent multi-aperture imaging system.}
	\label{fig:DAMIS}
\end{figure}

Our DMAIS consists of five long-focal cameras, each with a $4096\times3000$ pixel resolution and a $150mm$ lens focal length. The FoV of each camera is only about $5.4^\circ \times 4.0^\circ$. The structure diagram of the DMAIS is shown in \cref{fig:DAMIS}. The central camera, donated as $C_0$, observes the ground vertically downwards. The surrounding four cameras, labeled as $C_1$, $C_2$, $C_3$, and $C_4$, are positioned to face forward, backward, left, and right, respectively. The optical axis angle between camera $C_0$ and the other four cameras is approximately $45^\circ$. {This installation inclination angle is based on a comprehensive evaluation of the influence of pose estimation accuracy and observation angle. Our structural design is inspired by the Maltese-Cross model used in aerial multi-camera system\cite{Rupnik2015aerial}, where the inclination angle is typically designed between 30 and $45^\circ$. Therefore, we have selected a maximum of $45^\circ$ as the appropriate inclination angle.} Since there is no overlapping FoV among the five cameras, it is called a divergent multi-aperture imaging system. {The synchronization of image capture between cameras relies on GPS satellite signals as external hardware triggers. The system's time synchronization accuracy can reach the microsecond level.} The DMAIS is equipped with an onboard computing unit, which provides the GPS signal to simultaneously activate five cameras for image capture and storage. Additionally, it coordinates the functions of the pose estimation system.

The DMAIS is equipped on the flight platform to capture images of the ground scene, as shown in the left of \cref{fig:5cam}. Five long-focus cameras with a significant angular separation simultaneously capture images of the ground scene. As shown in the right of \cref{fig:5cam}, these five distinct, narrow FoV images can be merged to form a large $90^\circ \times 90^\circ$ FoV image. Each individual image has a high spatial resolution. Consequently, the DMAIS is equivalent to a single camera with both a large FoV and excellent spatial resolution. {Note that adding more cameras may further enhance system performance, but this is not guaranteed. If the additional cameras expand the overall FoV of the system, this will significantly enhance the spatial geometric constraints of the system. Therefore, these cameras will contribute to the improvement of pose estimation performance. However, the addition of more cameras also introduces challenges such as calibration complexity, computational load, and system weight. In this study, our goal is to find a balance between performance enhancement and system complexity.} Theoretically, the extraction accuracy and distribution of feature points directly determine the accuracy of pose estimation. Specifically, a higher spatial resolution of the image and a wider distribution range of feature points are positively associated with enhanced pose estimation accuracy. Thus, our DMAIS is expected to achieve more accurate pose estimation compared to using a single camera. 

Before pose estimation, we must perform geometric calibration of the DMAIS, which upgrades DMAIS from a simple image recording tool to a measuring instrument. The calibration process involves estimating both intrinsic and extrinsic parameters. {Intrinsic parameters include focal length, principal point, and distortion coefficients, which describe the camera's inherent geometric optical properties. Extrinsic parameters define the pose transformation of each camera with respect to the reference frame on the DMAIS.} Given known calibration parameters, the absolute pose estimation algorithm aims to determine the position and orientation of the DMAIS in the world frame. It takes only the 3D ground points in the world frame and their corresponding image 2D observations as input. The 3D world points can usually be obtained by matching the image with the pre-stored scene image. The DMAIS is a multi-camera system that requires a reasonable imaging model. This paper uses a generalized camera model to represent the DMAIS as a singular imaging system entity.


\section{Geometric Calibration}
\label{sec:calib}
Geometric calibration is the process of determining the mapping parameters between the 3D rays in space and 2D image pixels. In photogrammetry, achieving high-precision calibration is foundational to attaining accurate measurement. This paper proposes a calibration method for DMAIS based on the 3D calibration field. Our method first calibrates the intrinsic parameters of each long-focal camera separately. Then, we jointly calibrate the extrinsic parameters which refer to the relative pose between each camera and the reference frame.

\subsection{3D Calibration Field}

Each camera within DMAIS is equipped with a 150$mm$ long-focal lens, which has an extended observation range. Notably, there is no overlapping field of view between cameras. The calibration method based on planar target \cite{Zhang2000,Sturm1999,Li2013} is difficult to implement. This paper constructs a 3D calibration field to provide the pre-determined 3D scene structure for calibrating the DMAIS.

\begin{figure}[htpb]
	\centering
	\includegraphics[width=0.27\linewidth]{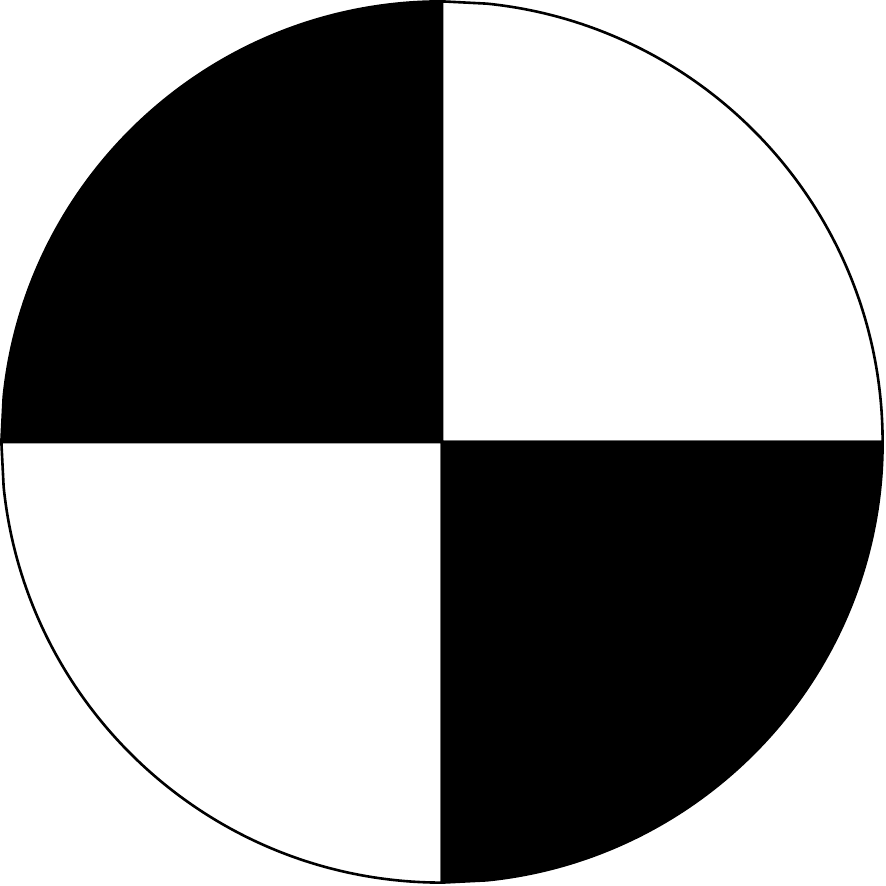}
	\quad
	\includegraphics[width=0.27\linewidth]{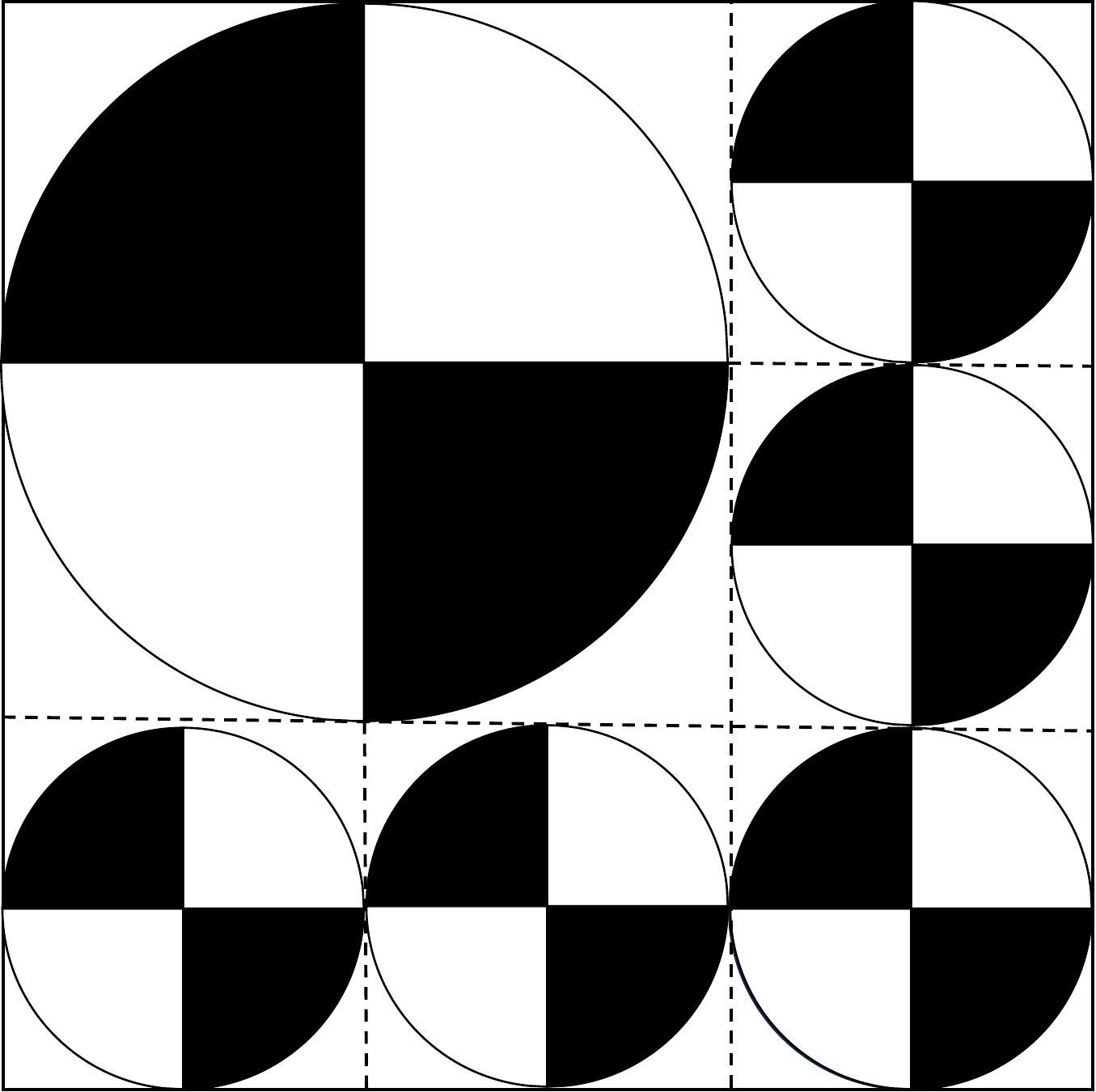}
	\quad
	\includegraphics[width=0.29\linewidth]{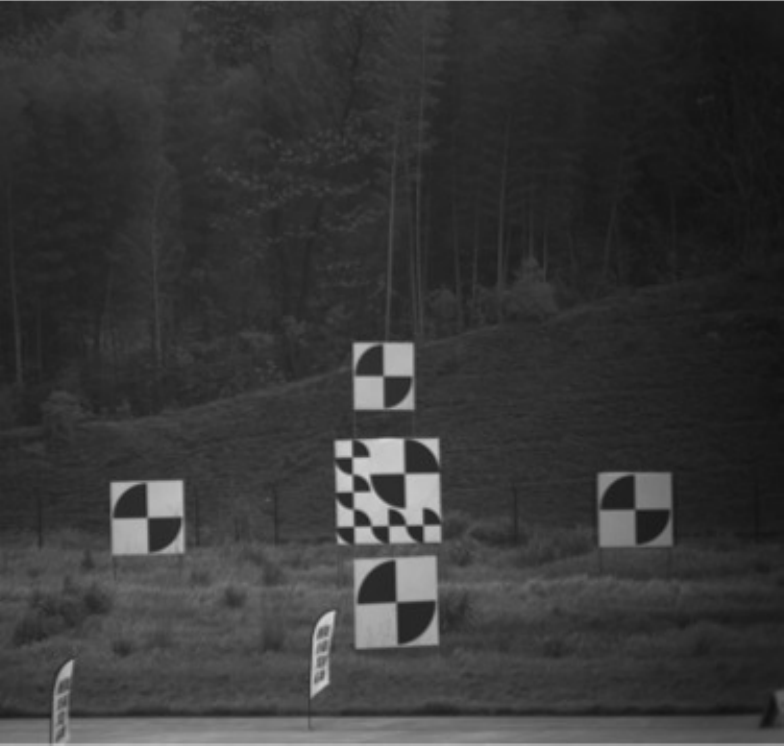}
	\caption{The control point composed of diagonal patterns.}
	\label{fig:diag}
\end{figure}

We select the simple diagonal pattern marker as the control point, as shown in \cref{fig:diag}. These markers can be accurately extracted with sub-pixel precision from images using a template matching method\cite{Wang2019}. To ensure visibility, individual markers are designed with size of $2m\times2m$. Additionally, we also customized a larger $3m\times3m$ marker, which contains six markers, as shown in the middle of \cref{fig:diag}. Markers are attached to the flat board to prevent any deformation, which could lower the calibration accuracy. 

The markers, serving as control points, are positioned near the ground, ranging from about 2 to 10 meters, as shown in right of \cref{fig:diag}. 
{ Placing control points at higher heights is impractical, especially in the wild.} We only construct local 3D calibration fields within the FoV of three horizontally positioned cameras to reduce the construction cost. {Not all markers are coplanar to prevent the degradation status.} The markers are placed perpendicular to the ground. The precise locations of the control point are measured using a combination of Real-Time Kinematic (RTK) positioning and total station measurements. 

{The measurement accuracy of control points is a key factor in determining the calibration accuracy of our imaging system. Our markers (i.e., control points) are arranged in an open, unobstructed outdoor environment, which provides ideal conditions for RTK measurements. The control points are positioned statically. The measurement accuracy achieved by the combination of RTK and total station can reach millimeter-level precision under static conditions. During the calibration process, we conduct multiple measurements of each control point's coordinates and used their average as the final result.} Each control point is assigned a unique number, corresponding to a unique 3D point. All the control points constitute the 3D calibration field of the known structure. Note that we have established a local coordinate system to determine the spatial coordinates of each control point, rather than directly using RTK measurement coordinates. The local coordinate system of the control points is regarded as the world frame.

\subsection{Intrinsic Calibration}
In this section, {intrinsic calibration is performed for each long-focal camera}. Here, considering that the distortion of long-focus lenses is relatively low, we model cameras using the well-known pinhole model with a single radial distortion coefficient. Given a 3D world point $\mathbf{M} = [X,Y,Z]^T$ and its corresponding 2D image point $\mathbf{m} = [u,v]$, the perspective projection process can be mathematically expressed as:
\begin{equation}
	s \widetilde{\mathbf{m}}=\mathbf{K}[\mathbf{R} \mid \mathbf{t}] \widetilde{\mathbf{M}} \text {, with } \mathbf{K}=\left[\begin{array}{ccc}
		f_{x} & \gamma & c_{x} \\
		0 & f_{y} & c_{y} \\
		0 & 0 & 1
	\end{array}\right] \text {, }
	\label{eq:sp=KRtP}
\end{equation}
where $s$ is an arbitrary scale factor denoting the depth of point, {$\mathbf{K}$ is the intrinsic matrix} and $(\mathbf{R}, \mathbf{t})$ is the camera pose relative to the world frame. $\mathbf{R}$ is a $3\times3$ rotation matrix and $\mathbf{t}$ is a $3\times1$ translation vector. The terms $\widetilde{\mathbf{m}} = [u,v,1]^T$ and $\widetilde{\mathbf{M}} = [X,Y,Z,1]^T$ represent the homogeneous form of $\mathbf{m}$ and $\mathbf{M}$, respectively. The intrinsic matrix $\mathbf{K}$ consists of five parameters: the focal length $(f_x,f_y)$, the principal point $(c_x,c_y)$ and skew factor $\gamma$. {The skew factor $\gamma$  is typically set to 0, and we ignore it following\cite{Sturm1999}.}

The linear projection model is insufficient for representing the actual imaging process. Specifically, the nonlinear distortion of lenses is essential for a comprehensive representation. The distortion levels of long-focal lenses are typically low, thus we only focus on the first term of radial distortion. Given a non-distortion projection point $(u,v)$, the distorted point $(\breve{u},\breve{v})$ is computed by:
\begin{equation}
	\left\{ {\begin{array}{*{20}{c}}
			{\breve{u}  = u + d\left( {u - {c_x}} \right)\left( {{x^2} + {y^2}} \right)}\\
			{\breve{v}  = v + d\left( {v - {c_y}} \right)\left( {{x^2} + {y^2}} \right)}
	\end{array}} \right. .
\end{equation}
Here, $d$ represents the radial distortion coefficient. The ideal normalized image coordinate $(x,y)$ can be computed as:
\begin{equation}
	\left\{ {\begin{array}{*{20}{c}}
			{x = \left( {u - {c_x}} \right)/{f_x}}\\
			{y = \left( {v - {c_y}} \right)/{f_y}}.
	\end{array}} \right.
\end{equation}

{The core of intrinsic parameter calibration is to compute the intrinsic matrix $\mathbf{K}$ and the distortion coefficients  $d$ for each camera.}
In this paper, we propose a unique method to obtain camera parameters via a 3D calibration field. First, the long-focal camera captures images of the arranged markers from multiple orientations. Then, the diagonal marker extraction is performed to find the corner coordinates of the diagonal markers in each image. The 2D-3D correspondences between 2D image corner points and 3D control points are then used to compute the initial estimation of the intrinsic parameters and camera poses. In initial estimation, we adopt some reasonable prior parameters: $d = 0$, $f_x = f_y = f$, $c_x = \frac{W}{2}$ and $c_y = \frac{H}{2}$ where $W$ and $H$ are the width and height of the image, respectively. Here, we can see that the initial intrinsic parameter to be estimated is only focal length $f$. The camera poses will be estimated as additional parameters.

To estimate the initial unknowns, we model the aforementioned problem as a PnPf problem, which refers to the perspective-n-point problem with unknown focal length. From Nakano's method \cite{Nakano2016} and the established 2D-3D correspondences, we simultaneously estimate the camera pose $(\mathbf{R}, \mathbf{t})$ and focal length$(f)$ for each image. Subsequently, we calculate the mean value of $f$ as the best hypothesis and re-estimate the pose for each image by solving the PnP problem. Up to now, we have obtained all initial guesses of unknowns. Finally, we jointly optimize the intrinsic parameters and camera poses using a nonlinear optimization step. We are given $n$ images, each of which contains $m$ points. The nonlinear optimization problem, which minimizes the sum of all re-projection errors, can be expressed as:
\begin{equation}
	\mathop{\min }\limits_\mathbf{X} \sum\limits_i^n {\sum\limits_j^m  {{{\left\| {\pi \left( {\mathbf{K},d,{\mathbf{R}_i},{\mathbf{t}_i},{\mathbf{M}_{ij}}} \right) - {\mathbf{m}_{ij}}} \right\|}^2}} },
	\label{eq:refine}
\end{equation}
where $\mathbf{X} = (\mathbf{K},d,\mathbf{R}_i, \mathbf{t}_i)$ contains all parameters to be optimized. $(\mathbf{R}_i, \mathbf{t}_i)$ is the pose of the $i$-th image in the world frame. $\mathbf{M}_{ij}$ represents the $j$-th 3D world point seen in the $i$-th image and corresponds to the 2D image point $\mathbf{m}_{ij}$. The projection function $\pi(\cdot)$ calculates the image coordinates of $\mathbf{M}_{ij}$ based on the camera parameters. {The objective of \cref{eq:refine} is to optimize $\mathbf{K}$ and $d$ such that the sum of distances between the computed projection points of 3D points $\pi \left( {\mathbf{K},d,{\mathbf{R}_i},{\mathbf{t}_i},{\mathbf{M}_{ij}}} \right)$  and their corresponding actual image points $\mathbf{m}_{ij}$ is minimized.} This nonlinear minimization problem \cref{eq:refine} can be iteratively solved using Levenberg-Marquardt algorithm\cite{more2016}, which is supported in several excellent frameworks\cite{Ceres,Lourakis2009}.

\subsection{Extrinsic Calibration}
Given the known intrinsic parameters of each camera, the extrinsic parameters of the DMAIS are then calibrated. The control points are arranged in three areas near the ground, and the DMAIS is placed horizontally to capture images of them.
The distribution of cameras in DMAIS has a limitation, as it only allows three horizontal cameras to simultaneously observe the ground-level control points, as shown in \cref{fig:calib1}. The remaining two cameras are oriented {toward the sky and ground}, respectively, and are thus unable to capture the arranged control points. Thus, we propose a new strategy to estimate the extrinsic parameters. Initially, the three horizontal cameras capture a set of images. Then, the DMAIS is rotated 90 degrees to allow the remaining two cameras to observe the control points. The cameras $C1-C4$ are rigidly attached to $C_0$ at 45 degrees. Following the rotation, only minor adjustments are required to ensure that the three new horizontal cameras observe the control points. After four such rotations, the DMAIS returns to its initial orientation, {acquiring of five sets of images.}

These five sets of images are used to {calibrate the extrinsic parameters accurately.} Firstly, the proposed method independently calculates the pose of each image in the world frame via image visual localization. Taking camera $C_0$ as the reference frame, the initial guess of extrinsic parameters is then calculated. Finally, the nonlinear optimization step jointly optimizes the extrinsic parameters and the poses of $C_0$ in the world frame.

\textbf{Image Visual Localization.}
The image visual localization takes images from the DMAIS and 3D control points as input, and outputs image poses in the world frame. For each image, we extract the pixel coordinates of the diagonal marker center and match them with the 3D control points individually. This allows us to establish correspondences between the 2D image and the 3D calibration field. Given 2D-3D correspondences, the PnP solvers \cite{Opnp2013,Lepetit2009} can be used to estimate image pose. {However, due to the narrow field of view of long-focal-length cameras, simple 2D-3D correspondences are insufficient to accurately constrain the image position.} A notable deviation exists between the estimated position and the actual position. 

To address this, the initial guess of the image position is measured in advance using RTK, which effectively constrains the image position and prevents the optimization process from falling into the local optimum. Compared with the large-scale 3D calibration field, {the measurement bias of the initial guess for image position is relatively minimal}. Furthermore, we use the DLT method\cite{Hartley2004} to estimate the image pose based on the known initial position and 2D-3D correspondences. Finally, we optimize the image pose using the standard bundle adjustment method\cite{Triggs2000}. The objective function is to minimize the re-projection error between the projected image points of the 3D control points and the actual image points.

\textbf{Initial Extrinsic Estimation.}
In this step, we estimate the initial extrinsic parameters of DMAIS using the image pose set obtained from the visual localization step. The five cameras are rigidly attached to the DMAIS without overlapping FoV. {Since camera $C_0$ can observe the control points during each rotation, we select it as the unified reference frame to connect with the other four surrounding cameras.}  We can express each image pose set as a function of the DMAIS pose and the extrinsic parameters. Here, the DMAIS pose represents the reference frame pose in the world frame. The extrinsic parameters denote the transformation matrix between other cameras and the reference frame. Let $\mathbf{P}_i^j$ represents the pose of camera $C_i (i = 0,...,4)$ in the world frame following the $j$-th rotation. The transformation matrix from the camera $C_i$ to the reference frame $C_0$ at the $j$-th rotation is given by:
\begin{equation}
	\mathbf{T}_{C_i}^j =  \mathbf{P}_0^j (\mathbf{P}_i^j)^{-1}.
\end{equation}
Assuming there are $k$ rotations, we then estimate the initial extrinsic parameters by averaging $\mathbf{T}_{C_i}^j (j=0,...,k)$:
\begin{equation}
	\mathbf{T}_{C_i} = \left[\begin{array}{cc}
		\mathbf{R}_{C_i} & \mathbf{t}_{C_i} \\
		\mathbf{0}_{1\times 3} & 1\end{array}\right] = \frac{1}{k} \sum \limits_{j=1}^k \mathbf{T}_{C_i}^j.
	\label{eq:avg_T}
\end{equation}
Note that \cref{eq:avg_T} is not a simple numerical average. We use the quaternion averaging method\cite{Markley2007} for rotation and basic numerical averaging for the translation vector. {The core of extrinsic parameter calibration is to compute the pose transformation $\mathbf{T}_{C_i}$ of each camera relative to the reference frame.}

\textbf{Nonlinear Optimization.}
The nonlinear optimization step is responsible for the final calibration accuracy. Based on the initial estimation, we jointly optimize the extrinsic parameters $\mathbf{T}_{C_i}$ ($i = 0,...,4$) and the poses of reference frame $\mathbf{P}_0^j$ ($j=0,...,k$). An optimization problem is formulated to minimize the sum of re-projection errors. The objective problem is expressed as:
\begin{equation} 
	\mathop {\min }\limits_{{\mathbf{P}_0^j},\mathbf{T}_{C_i}} \sum\limits_{i,j,m} {{{\left\| {\pi \left( {{\mathbf{K}_{C_i}},d_{C_i}, {\mathbf{P}_0^j},{\mathbf{T}_{C_i}},{\mathbf{M}_{m}}} \right) - {\mathbf{m}_{ijm}}} \right\|}^2}}.
	\label{eq:min}
\end{equation}
Here, $\mathbf{K}_{C_i}$ and $d_{C_i}$ represent the intrinsic parameters and distortion coefficient of camera ${C_i}$, $\mathbf{M}_m$ is the $m$-th control point, and $\mathbf{m}_{ijm}$ is the actual image point of $\mathbf{M}_m$ observed in camera $C_i$ with the reference frame pose $\mathbf{P}_{C_0}^j$. $\pi(\cdot)$ is a projection function that outputs the image coordinate of the control point $\mathbf{M}_m$ seen in camera $C_i$. {The objective of \cref{eq:min} is to optimize the extrinsic parameters $\mathbf{T}_{C_i}$ and the reference frame pose $\mathbf{P}_{C_0}^j$ such that the sum of distances between the computed projection points of 3D points $\pi \left( {{\mathbf{K}_{C_i}},d_{C_i}, {\mathbf{P}_0^j},{\mathbf{T}_{C_i}},{\mathbf{M}_{m}}} \right)$ and the corresponding actual image points $\mathbf{m}_{ijm}$ is minimized.} The Levenberg-Marquardt method\cite{more2016} is used to solve the nonlinear minimization problem.

\section{Absolute Pose Estimation}
\label{sec:pose}

Our pose estimation system aims to determine the pose of the DMAIS in the world frame using 2D images and 3D known structures. By matching features between the DMAIS images and the pre-stored structure, we can establish 2D-3D correspondences. In this section, we model the DMAIS as a generalized camera model and introduce an algorithm for absolute pose estimation. 

\subsection{Generalized Camera Model}
\begin{figure}[htbp]
	\centering
	\includegraphics[width=0.9\linewidth]{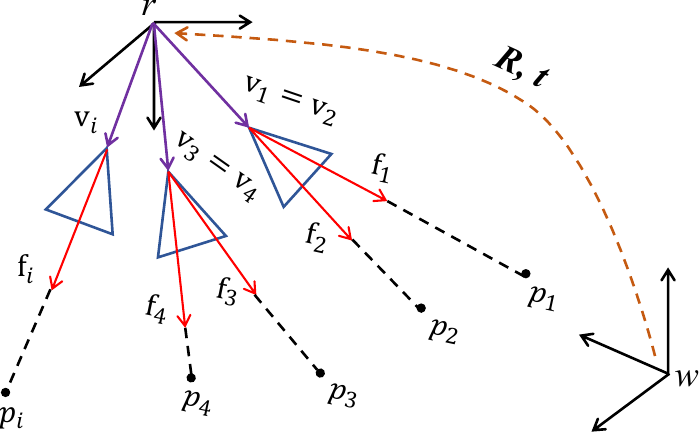}
	\caption{Projection diagram of generalized camera model.}
	\label{fig:gpnp}
\end{figure}

Inspired by Pless\cite{Pless2003}, we model the calibrated DMAIS as a generalized camera, which uses rays in space to represent image observations. The generalized camera model is briefly outlined below. Let $\mathbf{M}_i \in \mathbb{R}^3$ denote a 3D point in the world frame. As shown in \cref{fig:gpnp}, the observation of $\mathbf{M}_i$ in the generalized camera can be represented by a ray expressed as $\alpha_i \mathbf{f}_i + \mathbf{v}_i$. Here, $\mathbf{v}_i \in \mathbb{R}^3$ represents the relative position of camera $C_c (c = 0,...,4)$ in the reference frame, $\mathbf{f}_i \in \mathbb{R}^3$ is the unit direction vector of the ray, and $\alpha_i$ is the depth of $\mathbf{M}_i$. {Note that 3D points $\mathbf{p}_1$ and $\mathbf{p}_2$ are observed by the same camera (represented as a triangle), therefore $\mathbf{v}_1 = \mathbf{v}_2$. Similarly, points $\mathbf{p}_3$ and $\mathbf{p}_4$ are observed by the same camera, thus $\mathbf{v}_3 = \mathbf{v}_4$.} Given the intrinsic and extrinsic parameters of the DMAIS, the values of $\mathbf{v}_i$ and $\mathbf{f}_i$ can be calculated as follows:
\begin{align}
	\mathbf{v}_i  &= \mathbf{t}_{c} \\
	\mathbf{f}_i  &= \mathbf{R}_{c} \mathbf{K}_{c}^{-1} \widetilde{\mathbf{m}}_i
\end{align}
where $\mathbf{K}_{c}$ and $(\mathbf{R}_{c}, \mathbf{t}_{c})$ are the intrinsic and extrinsic parameters of camera $C_c$ respectively and $\widetilde{\mathbf{m}}_i$ is the homogeneous un-distorted image point corresponding to $\mathbf{M}_i$. 

In \cref{fig:gpnp}, we define $\mathbf{R}\in SO^3$ as the rotation matrix and $\mathbf{t} \in \mathbb{R}^3$ as the translation vector from the world frame $\{w\}$ to the reference frame in DMAIS $\{r\}$. Then, the projection equation of the generalized camera model can be expressed as
\begin{equation}
	\alpha_i \mathbf{f}_i + \mathbf{v}_i = \mathbf{R} \mathbf{M}_i + \mathbf{t}.
	\label{eq:projEq}
\end{equation}
Given multiple sets of inputs $\{\mathbf{v}_i, \mathbf{f}_i, \mathbf{M}_i\}$, the parameters to be estimated are the pose of DMAIS ($\mathbf{R}$ and $\mathbf{t}$) in the world frame. {Both rotation and translation have 3 DoF, resulting in a total of 6 DoF. Each 2D-3D point correspondence can constrain 2 DoF, so at least 3 non-collinear point pairs are required to achieve multi-camera pose estimation. Our absolute pose estimation algorithm also requires a minimum of 3 points to complete the solution, making it suitable for extremely low-texture environments. } We introduce the proposed pose estimation algorithm in detail below. 

\subsection{Pose Estimation Algorithm}
We approach the pose estimation problem of the generalized camera by framing it as a constrained optimization problem. Based on the collinear constraint, the space geometric error for each 3D point can be calculated as:
\begin{equation}
	e_i = ||[\mathbf{f}_i]_\times (\mathbf{R} \mathbf{M}_i + \mathbf{t}-\mathbf{v}_i)||^2,
	\label{eq:ei}
\end{equation}
where $[\mathbf{f}_i]_\times$ is the antisymmetric matrix of $\mathbf{f}_i$. The space geometric error comes from the simple fact that the cross-product of two vectors results in a zero vector when they share the same direction. The depth factor $\alpha$ is efficiently eliminated in \cref{eq:ei}. Given $n$ 3D points and corresponding observations, we calculate the optimal pose of the DMAIS by minimizing the sum of errors. {We model absolute pose estimation as a constrained optimization problem.} The objective function is mathematically expressed as
\begin{equation}
	\left\{
	\begin{aligned}
		\min _{\mathbf{R}, \mathbf{t}}& \sum_{i=1}^{n}\left\|[\mathbf{f}_i]_\times \left(\mathbf{R} \mathbf{M}_{i}+\mathbf{t}-\mathbf{v}_{i}\right)\right\|^{2} \\
		\mathrm{s.t.}& \mathbf{R}^T\mathbf{R}=\mathbf{I},\quad \det(\mathbf{R})=1.
	\end{aligned}
	\right.
	\label{eq:objfun}
\end{equation}
The constraints in \cref{eq:objfun} arise from the fact that $\mathbf{R}$ is an orthogonal matrix. {The objective of \cref{eq:objfun} is to optimize the pose of the multi-aperture imaging system $(\mathbf{R}, \mathbf{t})$ such that the sum of space geometric errors is minimized.} Let $\mathbf{r}$ be a 9D vector formed by vectorizing the rotation matrix $\mathbf{R}$, where $\mathbf{r} = [R_{1},R_{2},R_{3},...,R_{9}]^T$ and $R_{1},R_{2},...,R_{9}$ are the elements of $\mathbf{R}$. For each 3D point $\mathbf{M}_i$, we define a corresponding $3\times9$ matrix $\mathbf{Y}_i$ such that $\mathbf{R} \mathbf{M}_i = \mathbf{Y}_i \mathbf{r}$. The matrix $\mathbf{Y}_i$ can be represented as
\begin{equation}
	\mathbf{Y}_i = \left[\begin{array}{lll}
		\mathbf{M}_{i}^{T} & & \\
		& \mathbf{M}_{i}^{T} & \\
		& & \mathbf{M}_{i}^{T}
	\end{array}\right].
\end{equation}
Then, the objective function is rewritten as
\begin{equation}
	\left\{
	\begin{aligned}
		\min _{\mathbf{r}, \mathbf{t}}& \left\| \mathbf{A}\mathbf{r}+\mathbf{B}\mathbf{t}-\mathbf{D}\right\|^2 \\
		\mathrm{s.t.}& \mathbf{R}^T\mathbf{R}=\mathbf{I},\quad \det(\mathbf{R})=1.
	\end{aligned}
	\right.
	\label{eq:objfun2}
\end{equation}
where
\begin{equation}
	\scriptsize
	\mathbf{A}_{3n\times9} \! = \! \left[\!\!\!\begin{array}{c}
		{\left[\mathbf{f}_1\right]_{\times} \mathbf{Y}_1} \\
		\vdots \\
		{\left[\mathbf{f}_n\right]_{\times} \mathbf{Y}_n}
	\end{array}\!\!\!\right],
	\mathbf{B}_{3n\times3}\! = \! \left[\!\!\!\begin{array}{c}
		{\left[\mathbf{f}_1\right]_{\times}} \\
		\vdots \\
		{\left[\mathbf{f}_n\right]_{\times}}
	\end{array}\!\!\!\right], \\
	\mathbf{D}_{3n\times1}\! = \!\left[\!\!\!\begin{array}{c}
		{\left[\mathbf{f}_1\right]_{\times} \mathbf{v}_1} \\
		\vdots \\
		{\left[\mathbf{f}_n\right]_{\times} \mathbf{v}_n}
	\end{array}\!\!\!\right].
\end{equation}

The translation vector $\mathbf{t}$ is an unconstrained variable. Suppose $\mathbf{R}$ is known, \cref{eq:objfun2} becomes a linear least squares problem with respect to $\mathbf{t}$. Let the derivative of $\left\| \mathbf{A}\mathbf{r}+\mathbf{B}\mathbf{t}-\mathbf{D}\right\|^2$ with respect to $\mathbf{t}$ be equal to $\mathbf{0}$, we obtain the least square solution of $\mathbf{t}$:
\begin{equation}
	\mathbf{t} = -\mathbf{B}^{+} (\mathbf{A}\mathbf{r}-\mathbf{D}), 
	\label{eq:t}
\end{equation}
where $\mathbf{B}^{+}$ is the pseudo-inverse matrix of $\mathbf{B}$. This expression shows $\mathbf{t}$ as a linear function of $\mathbf{R}$. Substituting \cref{eq:t} into \cref{eq:objfun2}, we get the final form of objective function
\begin{equation}
	\left\{
	\begin{aligned}
		\min_{\mathbf{r}}& (\mathbf{A}\mathbf{r}-\mathbf{D})^T \mathbf{G} (\mathbf{A}\mathbf{r}-\mathbf{D}) \\
		\mathrm{s.t.}& \mathbf{R}^T\mathbf{R}=\mathbf{I},\quad \det(\mathbf{R})=1,
	\end{aligned}
	\right.
	\label{eq:objfun3}
\end{equation}
where $\mathbf{G} = (\mathbf{I}-\mathbf{B}\mathbf{B}^{+})^T(\mathbf{I}-\mathbf{B}\mathbf{B}^{+})$. To address this equality-constrained optimization problem, the Lagrangian multiplier method is applied. The Lagrange function of \cref{eq:objfun3} can be expressed as
\begin{equation}
	\begin{aligned}
		L(\mathbf{R},\mathbf{S},\lambda)=&\frac12(\mathbf{A}\mathbf{r}-\mathbf{D})^T \mathbf{G} (\mathbf{A}\mathbf{r}-\mathbf{D}) \\ 
		&-\frac12\mathrm{trace}\left(\mathbf{S}(\mathbf{R}^T\mathbf{R}-\mathbf{I})\right)\\
		&-\lambda\left(\det\left(\mathbf{R}\right)-1\right),
	\end{aligned}
\end{equation}
where $\lambda$ is a Lagrange multiplier and $\mathbf{S}$ is a symmetric matrix of Lagrange multipliers. Then, the optimality conditions are expressed as follows:
\begin{flalign}
	\frac{\partial L}{\partial \mathbf{R}} &=\emph{mat} \left(\mathbf{A}^T\mathbf{G}\mathbf{A}\mathbf{r}-\mathbf{A}^T\mathbf{G}\mathbf{D}\right)-\mathbf{R}\mathbf{S}-\lambda\mathbf{R}=\mathbf{0}, \label{eq:L_R}\\
	\frac{\partial L}{\partial\mathbf{S}}&=\mathbf{R}^T\mathbf{R}-\mathbf{I}=\mathbf{0}, \\
	\frac{\partial L}{\partial\lambda}&=\det(\mathbf{R})-1=0, &
\end{flalign}
where $\emph{mat}(\cdot)$ is the operation of reshaping a 9D vector to a $3 \times 3$ matrix. For convenience, let us define a function $\mathcal{M}(\mathbf{r}) = \text{mat} \left(\mathbf{A}^T\mathbf{G}\mathbf{A}\mathbf{r}-\mathbf{A}^T\mathbf{G}\mathbf{D}\right)$. From \cref{eq:L_R}, we get two equation 
\begin{flalign}
	\mathbf{R}^T\mathcal{M}(\mathbf{r}) &= \mathbf{S} + \lambda \mathbf{I}, \\
	\mathcal{M}(\mathbf{r})\mathbf{R}^T &= \mathbf{R}(\mathbf{S} + \lambda \mathbf{I})\mathbf{R}^T.&
\end{flalign}
Note that $(\mathbf{S} + \lambda\mathbf{I})$ and $\mathbf{R}(\mathbf{S} + \lambda\mathbf{I})\mathbf{R}^T$ are symmetric matrices, we have the following two constraints
\begin{flalign}
	\mathbf{E}&=\mathbf{R}^T\mathcal{M}(\mathbf{r})-\mathcal{M}(\mathbf{r})^T\mathbf{R} = \mathbf{0}, \label{eq:E}\\
	\mathbf{F}&=\mathcal{M}(\mathbf{r})\mathbf{R}^T-\mathbf{R}\mathcal{M}(\mathbf{r})^T = \mathbf{0}. \label{eq:F}&
\end{flalign}
Let $E_{ij}$ and $F_{ij}$ denote the elements of $i$-th row and $j$-th column of $\mathbf{E}$ and $\mathbf{F}$, respectively. From \cref{eq:E} and \cref{eq:F}, we have 6 polynomial equations
\begin{equation}
	\begin{aligned}
		E_{12} = 0, \quad E_{13} = 0,\quad E_{23} = 0, \\
		F_{12} = 0,\quad F_{13} = 0,\quad F_{23} = 0.
	\end{aligned}
	\label{eq:EF6}
\end{equation}

The unknown in polynomial equations \cref{eq:EF6} is a rotation matrix. The 9-parameter matrix representation for the rotation is redundant as rotation typically has only 3 degrees of freedom. To address this redundancy, we choose a more compact Cayley parameterization method, which uses only 3 parameters to represent the rotation matrix. The rotation matrix can be expressed as
\begin{equation}
	\small
	\begin{aligned}
		&\mathbf{R}=\frac1{1+q_x^2+q_y^2+q_z^2} \cdot \\&\begin{bmatrix}1+q_x^2-q_y^2-q_z^2&2q_xq_y-2q_z&2q_y+2q_xq_z\\2q_xq_y+2q_z&1-q_x^2+q_y^2-q_z^2&2q_yq_z-2q_x\\2q_xq_z-2q_y&2q_x+2q_yq_z&1-q_x^2-q_y^2+q_z^2\end{bmatrix}
	\end{aligned}
\end{equation}
The Cayley parameterization method reduces the number of unknowns from 9 to 3. {Given correspondences between 3D points $\mathbf{M}_i (i = 1,...,n)$ and image observations $(\mathbf{f}_i, \mathbf{v}_i) (i = 1,...,n)$, we get 6 polynomial equations in \cref{eq:EF6} for three unknowns $\{q_x, q_y,q_z\}$}.

The polynomial equations \cref{eq:EF6} can be solved using the Gr{\"o}bner basis method, a widely used technique in visual geometry. The Gr{\"o}bner basis is a special set of polynomial equations. {These equations share the same solutions as the original but are easier to solve.} Thus, we directly use a Gr{\"o}bner basis solver generator proposed by Kukelova et al.\cite{Kukelova2008} The generator outputs code to construct the elimination template and the action matrix. With this code, the solution of polynomial equations $\{q_x, q_y,q_z\}$ can be calculated. Once the rotation matrix parameters $\mathbf{R}$ are obtained, we substitute them into \cref{eq:t} to calculate the optimal translation $\mathbf{t}$. So far, the pose of DMAIS in the world frame is completely estimated.

\section{Experimental Result}
\label{sec:result}
\subsection{Geometric Calibration Results}

\begin{figure*}[htbp]
	\centering
	\subfloat[]{
		\includegraphics[width=0.24\linewidth]{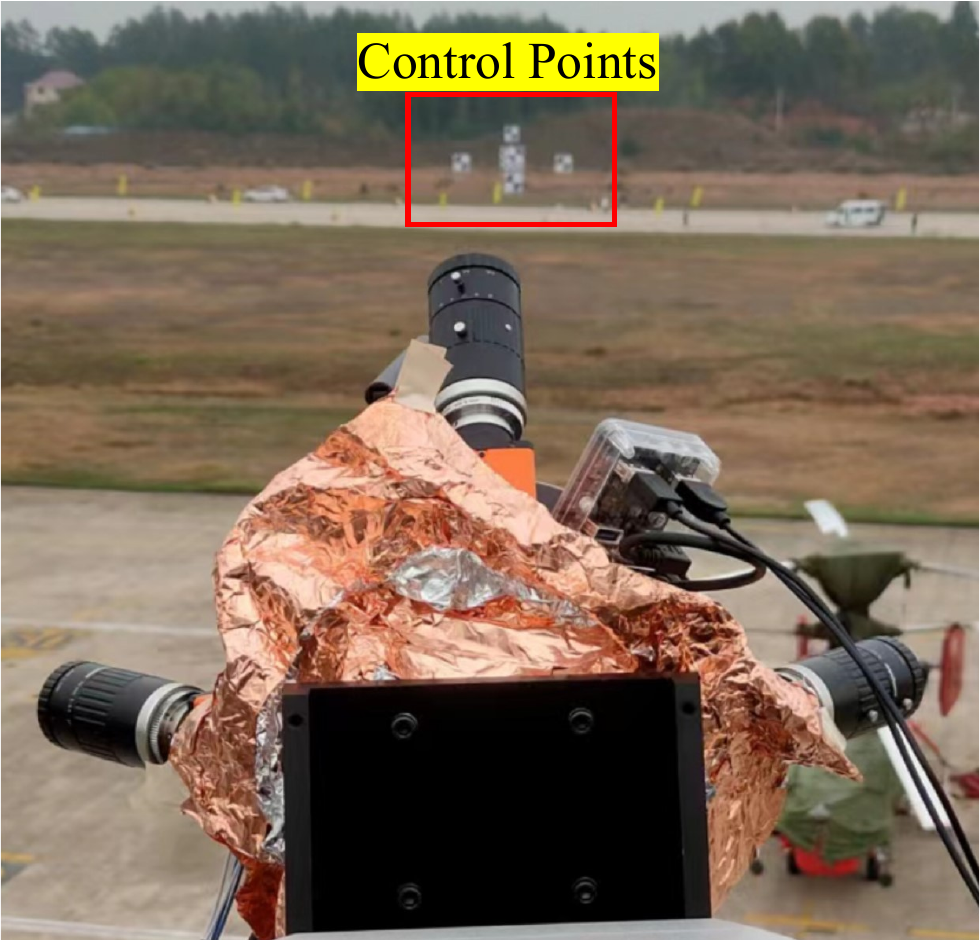} 
		\label{fig:calib1}
	}
	\subfloat[]{
		\includegraphics[width=0.24\linewidth]{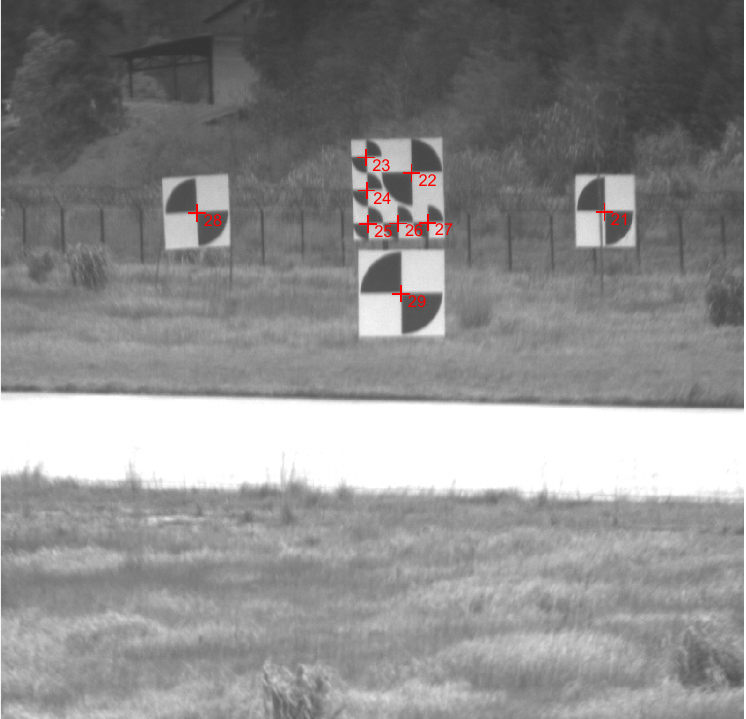}
		\label{fig:calib2}
	}
	\subfloat[]{
		\includegraphics[width=0.24\linewidth]{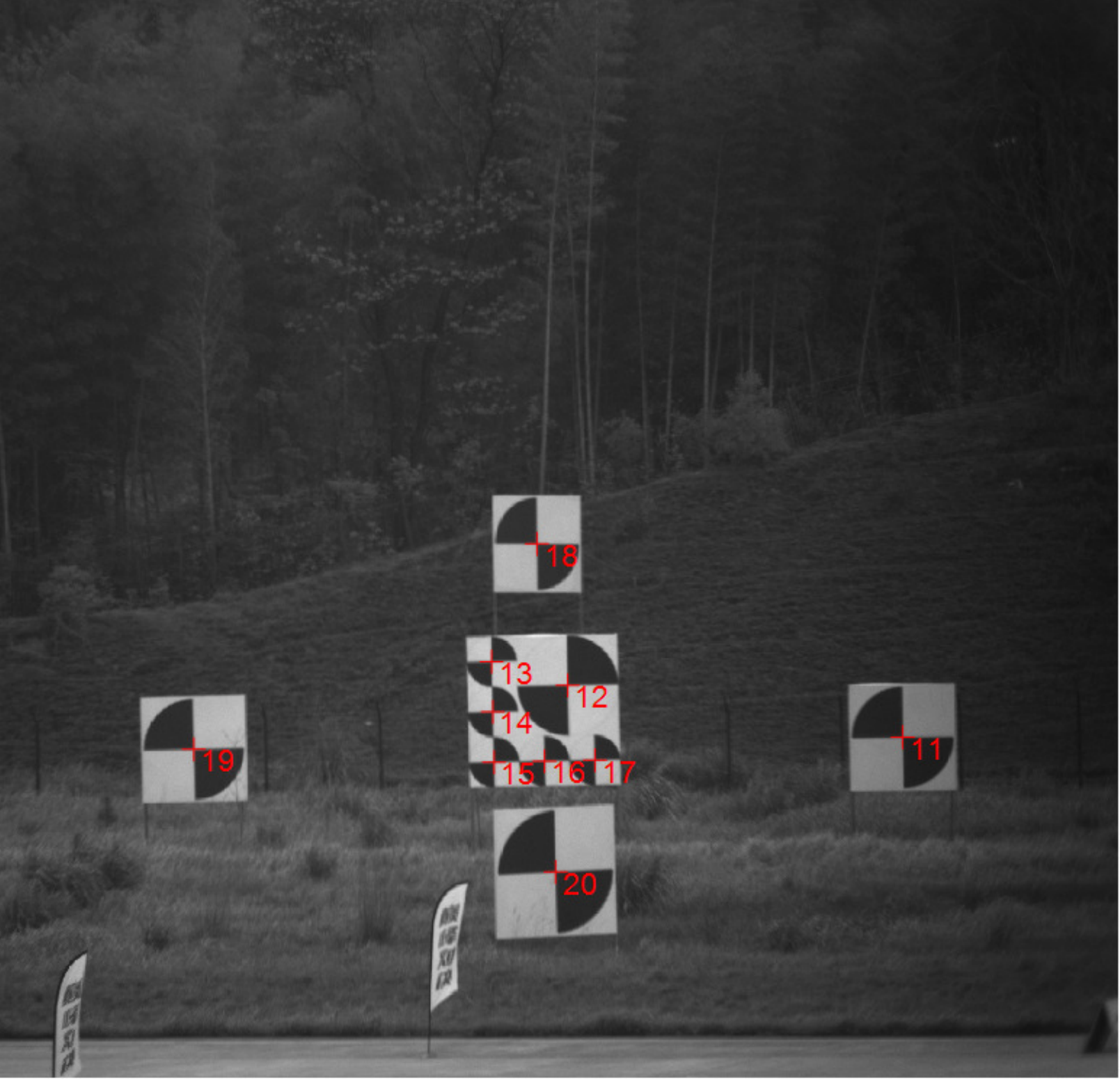} 
		\label{fig:calib3}
	}
	\subfloat[]{
		\includegraphics[width=0.24\linewidth]{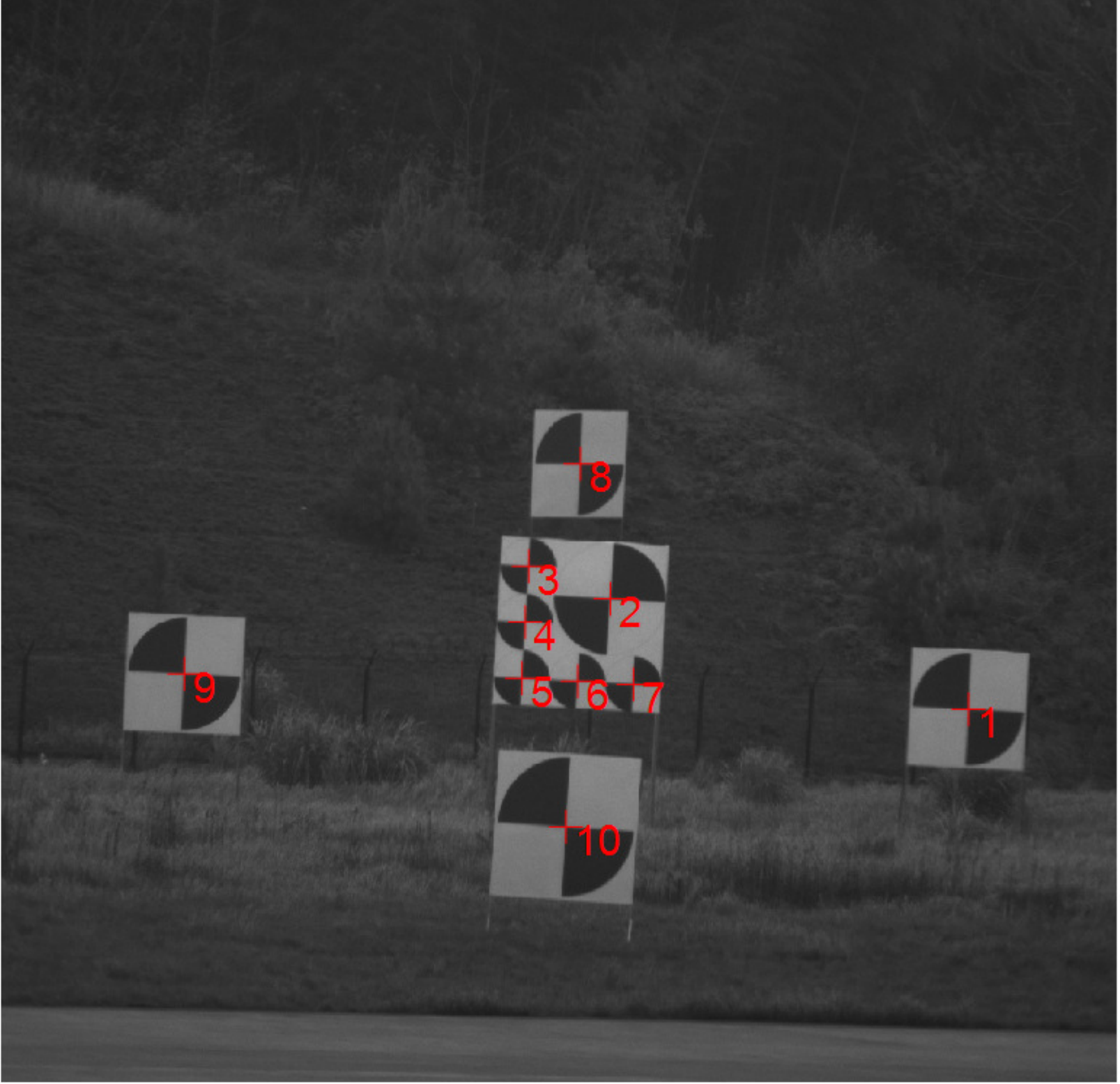} 
		\label{fig:calib4}
	}
	\caption{The examples of DMAIS and 3d calibration field. (a) The divergent multi-aperture imaging system to be calibrated; (b), (c), (d) The distribution of control points and the results of marker extraction.}
	\label{fig:calib}
\end{figure*}

\begin{table*}[htbp]
	\centering
	\caption{{Comparison of the intrinsic parameters calibration result.}}
	\resizebox{\linewidth}{!}{
		\begin{tabular}{clccccccc}
			\hline
			\multicolumn{2}{c}{Parameters}& $f_x(px)$ & $f_y(px)$ & $c_x(px)$ & $c_y(px)$ & $d$ & $\varepsilon_{calib}(px)$ & $\varepsilon_{eval}(px)$\\
			\hline
			\multirow{4}{*}{$C_0$} & Bouguet & 45466.2 & 45481.9 & 1941.0 & 1421.9 & 2.9600 & 0.2178 & 0.4315 \\
			& OpenCV & 46820.5 & 16809.6 & 1994.8 & 1777.1 & 12.8941 & 1.9386 & 1.8033 \\
			& BabelCalib & 45466.3 & 45473.0 & 1956.7 & 1413.4 & 3.0423 & 0.1872 & \underline{0.4231} \\
			& Ours & 45553.0 & 45569.7 & 2047.7 & 1500.0 & 3.0715 & 0.2726 & \textbf{0.3783} \\
			\hline
			\multirow{4}{*}{$C_1$} & Bouguet & 44262.2 & 42425.3 & 2225.6 & 1504.2 & 2.7286 & 0.3369 & 0.4510 \\
			& OpenCV & 45152.4 & 45392.3 & 2406.8 & 1274.1 & 5.1593 & 0.8342 & 1.0207 \\
			& BabelCalib & 44704.4 & 44684.9 & 2092.2 & 1282.7 & 1.8908 & 0.8998 & \underline{0.4356} \\
			& Ours & 44922.5 & 44963.4 & 2048.1 & 1498.9 & 2.0227 & 0.4534 & \textbf{0.4101} \\
			\hline
			\multirow{4}{*}{$C_2$} & Bouguet & 44222.3 & 44205.2 & 1894.1 & 1559.6 & 2.7520 & 0.3105 & \textbf{0.1822} \\
			& OpenCV & 45149.9 & 45403.6 & 2125.5 & 1255.47 & 4.8786 & 0.6642 & 0.5830 \\
			& BabelCalib & 44178.0 & 44143.9 & 2065.7 & 1387.7 & 2.8792 & 0.8975 & 0.2351 \\
			& Ours & 44440.4 & 44411.0 & 2048.3 & 1500.7 & 2.2361 & 0.3662 & \underline{0.2164} \\
			\hline
			\multirow{4}{*}{$C_3$} & Bouguet & 43813.9 & 43751.5 & 2047.5 & 1499.5 & 2.8741 & 0.2989 & \underline{0.2920} \\
			& OpenCV & 45208.3 & 45375.6 & 2067.1 & 963.2 & 2.6890 & 0.6551 & 0.5970 \\
			& BabelCalib & 42938.3 & 44264.7 & 2117.2 & 1546.1 & 2.6942 & 0.5353 & 0.3855 \\
			& Ours & 43737.5 & 43655.6 & 2047.6 & 1501.2 & 2.2207 & 0.3077 & \textbf{0.2878} \\
			\hline
			\multirow{4}{*}{$C_4$} & Bouguet & 44211.4 & 44179.0 & 2200.3 & 1455.9 & 2.6552 & 0.3160 & \underline{0.6375} \\
			& OpenCV & 45173.0 & 45373.4 & 2218.5 & 1050.1 & 4.4547 & 0.8390 & 1.1196 \\
			& BabelCalib & 44178.4 & 44157.1 & 2265.6 & 1290.7 & 2.1500 & 0.3375 & 0.6808 \\
			& Ours & 44660.0 & 44620.9 & 2048.2 & 1500.6 & 1.8864 & 0.4090 & \textbf{0.6303} \\
			\hline
		\end{tabular}
	}
	\normalsize
	\label{tab:intrinsics}
\end{table*}
Before the pose estimation task, we perform effective geometric calibration of the DMAIS to obtain intrinsic and extrinsic parameters. Firstly, a 3D calibration field is set up to provide the necessary input. The DMAIS is placed on the rooftop of a building, and we arrange control points within the FoV of the three horizontal cameras. The DMAIS setup and distribution of control points can be seen in \cref{fig:calib}. The DMAIS is rotated multiple times to ensure each camera captures the control points. Subsequently, the centers of diagonal markers in the images are manually extracted using the marker extraction method\cite{Wang2019,Tao2024}. The results of the marker extraction process are depicted in \cref{fig:calib2}, \cref{fig:calib3} and \cref{fig:calib4}.

In the initial experiment, we determine the intrinsic parameters of each camera. To quantitatively evaluate the results, we categorize all images into two groups: calibration images and evaluation images. The calibration images are used to estimate camera intrinsic parameters with $\varepsilon_{calib}$ representing the calibration re-projection error. Subsequently, using the obtained intrinsic parameters, we calculate the pose of evaluation images and report the re-projection error $\varepsilon_{eval}$. The units of $\varepsilon_{calib}$ and $\varepsilon_{eval}$ are pixels, denoted as $px$. {To more comprehensively evaluate the performance of our method, we have included comparisons with the following methods:}
\begin{itemize}
	\item Bought \cite{Bouguet2004}: The Bouguet toolbox is a widely used camera calibration tool and is also well-suited for calibrating 3D targets.
	\item {OpenCV \cite{Opencv2000}: OpenCV is a widely used open-source library for computer vision that includes mature calibration solutions.}
	\item {BabelCalib \cite{Lochman2021babelcalib}: BabelCalib is a universal camera calibration method that uses a back-projection model to decouple the calibration task of camera models into simpler regression tasks. }
\end{itemize}
{\cref{tab:intrinsics} presents the calibration results of intrinsic parameters and the evaluation error for each camera using the comparison methods. The smallest evaluation errors are bolded, and the second smallest are underlined. Obviously, our calibration results demonstrate the smallest evaluation errors, outperforming the other methods. Typically, the principal point $(c_x, c_y)$ is difficult to constrain when calibrating a long-focal camera. In this case, our method effectively constrains the principal point at the center of the image, while other methods exhibit larger offsets. This advantage arises from our reasonable assumptions and the initial values computed using the PnPf method\cite{Nakano2016}, which effectively prevents the objective function from getting trapped in local optima. Compared with our method, Bouguet's Toolbox\cite{Bouguet2004} tends to overfit, resulting in lower calibration re-projection error $\varepsilon_{calib}$. However, our method demonstrates superior accuracy in the evaluation images, indicating a better calibration result.
}
\begin{table}[tbp]
	\centering
	\caption{The calibration results of relative rotation with the cases of known and unknown initial position.}
	\resizebox{\linewidth}{!}{
		\begin{tabular}{cccc}
			\toprule
			Cameras & $\mathbf{R}_{known}(deg)$ & $\mathbf{R}_{unknown}(deg)$ & $\varepsilon_{\mathbf{R}}(deg)$ \\ 
			\midrule
			$C_1$ & (-179.811, -0.152, 46.233) & (-179.871, -0.080, 46.271) & 0.101 \\
			$C_2$ & (-90.048, -1.170, 44.758) & (-90.043, -1.117, 44.754) & 0.054 \\
			$C_3$ & (0.150, -0.047, 44.038) & (0.220, -0.149, 44.051) & 0.125 \\
			$C_4$ & (90.140, 1.180, 45.615) & (90.116, 1.168, 45.253) & 0.363 \\
			\bottomrule
		\end{tabular}
	}
	\label{tab:extrinsics_R}
\end{table}

\begin{table}[tbp]
	\centering
	\caption{The calibration results of relative translation with the cases of known and unknown initial position.}
	\resizebox{\linewidth}{!}{
		\begin{tabular}{cccc}
			\toprule
			Cameras & $\mathbf{t}_{known}(m)$ & $\mathbf{t}_{unknown}(m)$ & $\varepsilon_{\mathbf{t}}(m)$ \\ 
			\midrule
			$C_1$ & (-0.002, 0.102, -0.086) & (0.378, -2.765, -1.933) & 3.432 \\
			$C_2$ & (-0.113, -0.012, -0.100) & (1.111, 0.032, -0.777) & 1.400 \\
			$C_3$ & (0.000, -0.143, -0.097) & (-0.096, -0.464, 1.235) & 1.373 \\
			$C_4$ & (0.122, -0.010, -0.100) & (0.944, -0.167, 1.403) & 1.720 \\
			\bottomrule
		\end{tabular}
	}
	
	\label{tab:extrinsics_t}
\end{table}

\begin{figure}[tbp]
	\centering
	\subfloat[]{
		\includegraphics[width=0.45\linewidth]{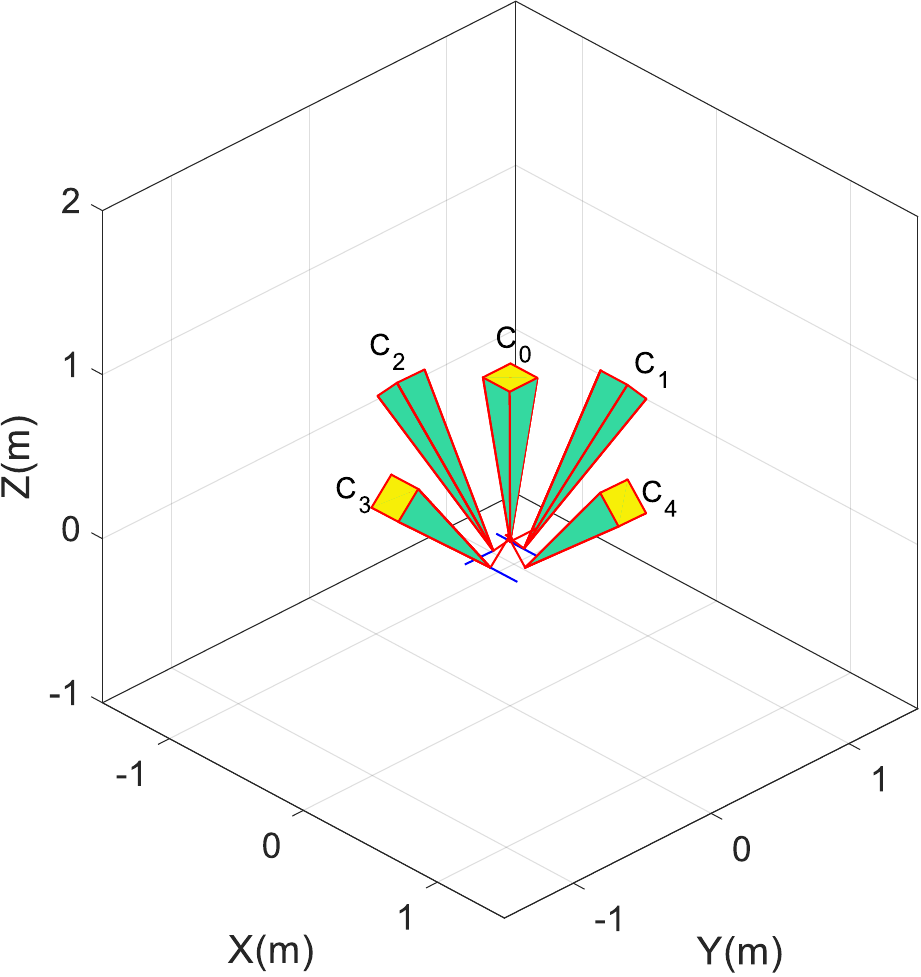} 
		\label{fig:known}
	}
	\subfloat[]{
		\includegraphics[width=0.47\linewidth]{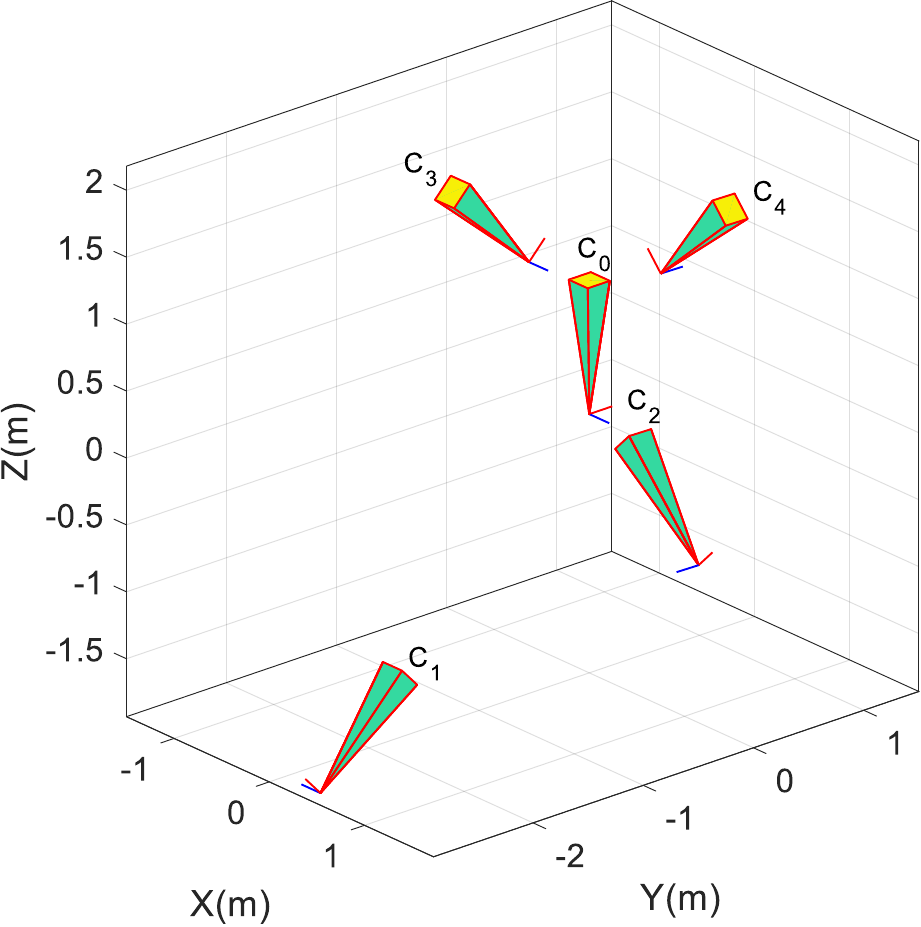} 
		\label{fig:unknown}
	}
	\caption{The relative distribution of five cameras in DMAIS. (a) The initial position is known. (b)The initial position is unknown.}
	\label{fig:extrinsics}
\end{figure}

After determining the intrinsic parameters of each camera, we then calibrate the extrinsic parameters of DMAIS. The local frame of camera $C_0$ serves as the reference frame. To effectively constrain the relative translation, we use RTK to measure the position of each camera as the initial value in the visual localization process. In this experiment, we compare the extrinsic parameters results in two cases: known and unknown initial positions.

\cref{tab:extrinsics_R} shows the relative rotation results of the two cases. We use the Euler angle to represent relative rotation in $ZYX$ order. The difference between two rotations is denoted as $\epsilon_\mathbf{R}$, which is calculated as the distance between two Euler angles. The relative translation results are shown in \cref{tab:extrinsics_t}. $\epsilon_\mathbf{t}$ represents the distance between two translation vectors. We can see from \cref{tab:extrinsics_R} that the difference between rotations is very small, with a maximum variance of only 0.363 degrees. However, there are significant differences in translations, with a minimum distance of 1.3 meters and a maximum of 3.4 meters. {As shown in \cref{tab:extrinsics_t}, the calculated relative translation between cameras significantly deviates from the actual values when no initial position is provided.} The narrow field of view enables accurate estimation of camera orientation but shows the challenge in constraining position.

For a more intuitive representation of the extrinsic calibration results, we show the relative distribution of five cameras in \cref{fig:extrinsics}. The results in \cref{fig:known} are consistent with the distribution of cameras in the actual DMAIS, while the results in \cref{fig:unknown} have a significant deviation from the actual. Finally, we choose the extrinsic parameters shown in Fig. \ref{fig:known} as the final calibration results. {These experimental findings provide significant practical support for the calibration of long-focus multi-camera systems.}

\subsection{Pose Estimation Results}

\begin{figure}[tbp]
	\centering
	\includegraphics[width=0.9\linewidth]{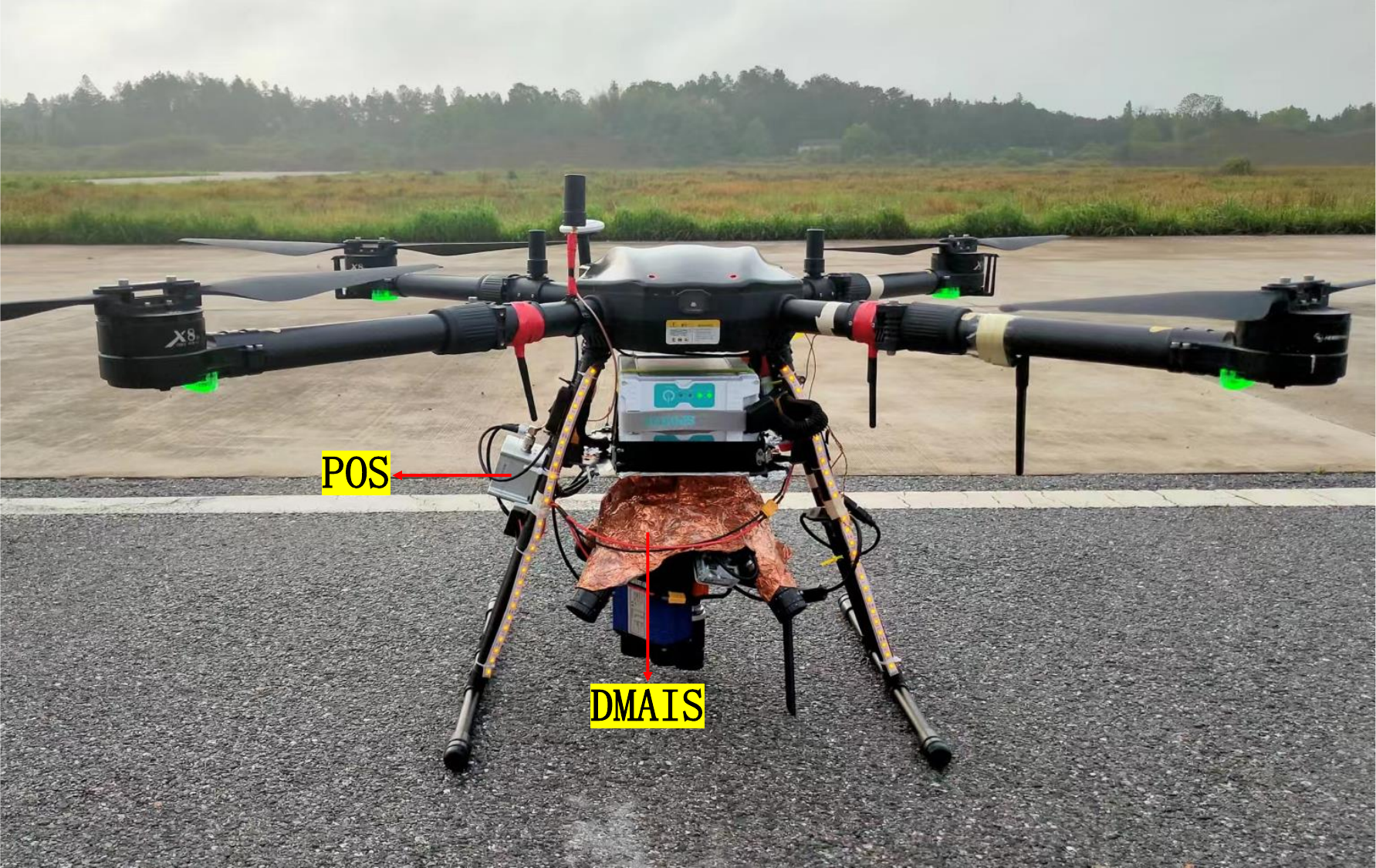}
	\caption{The UAV is equipped with DMAIS and POS for experiments.}
	\label{fig:uav}	
\end{figure}

\begin{figure*}[htbp]
	\centering
	\includegraphics[width=0.25\linewidth]{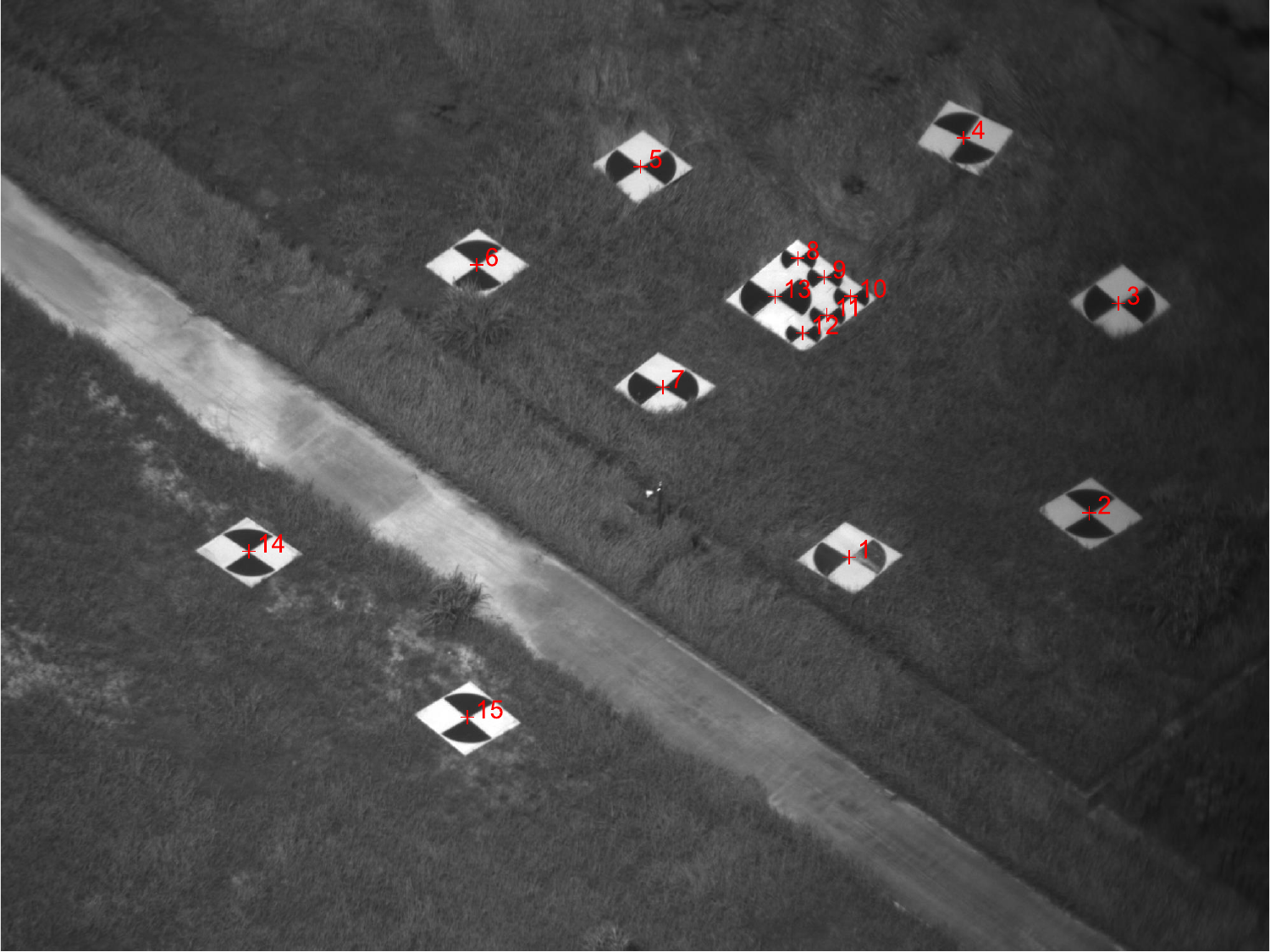} 
	\includegraphics[width=0.25\linewidth]{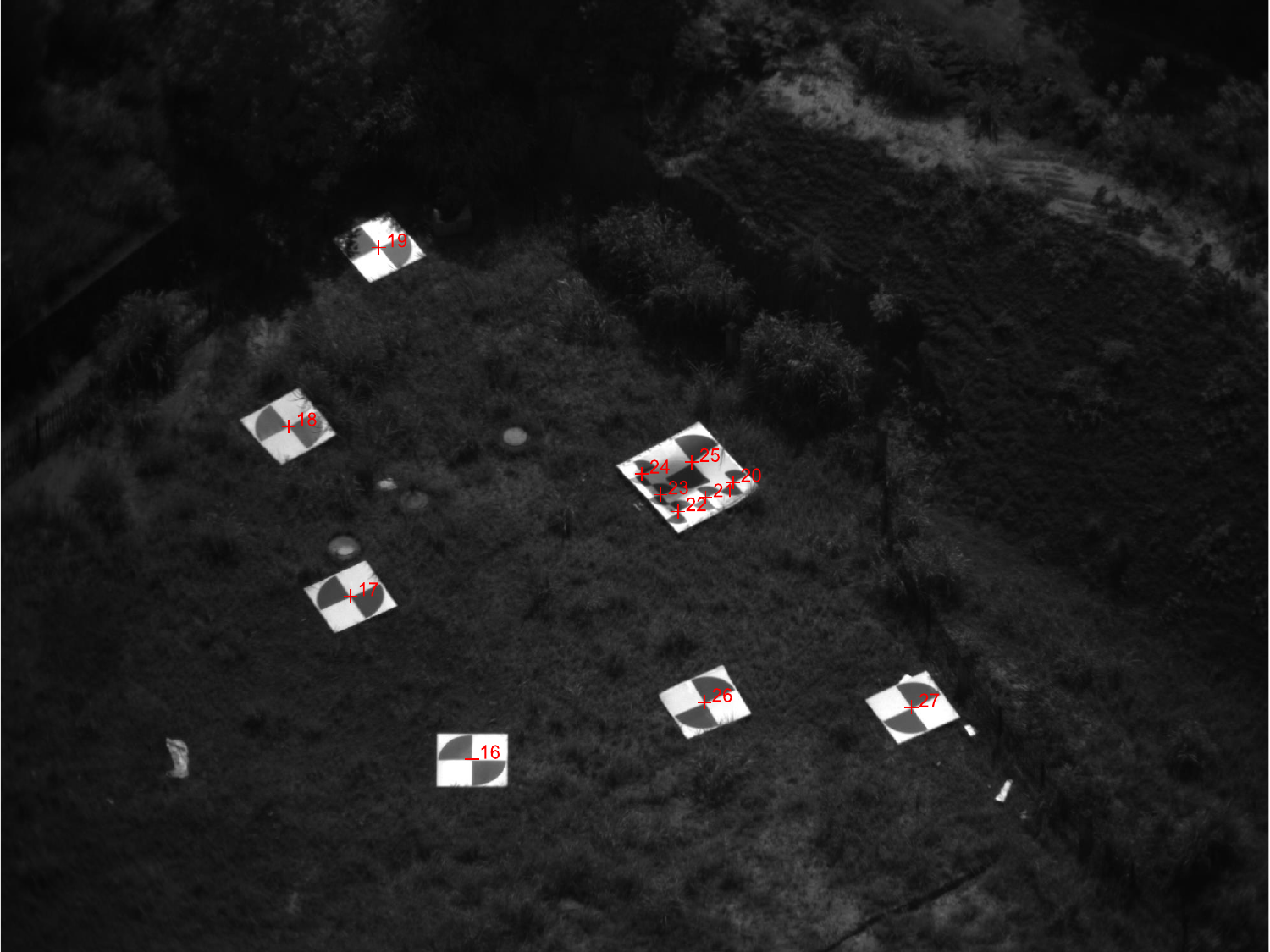}
	\includegraphics[width=0.25\linewidth]{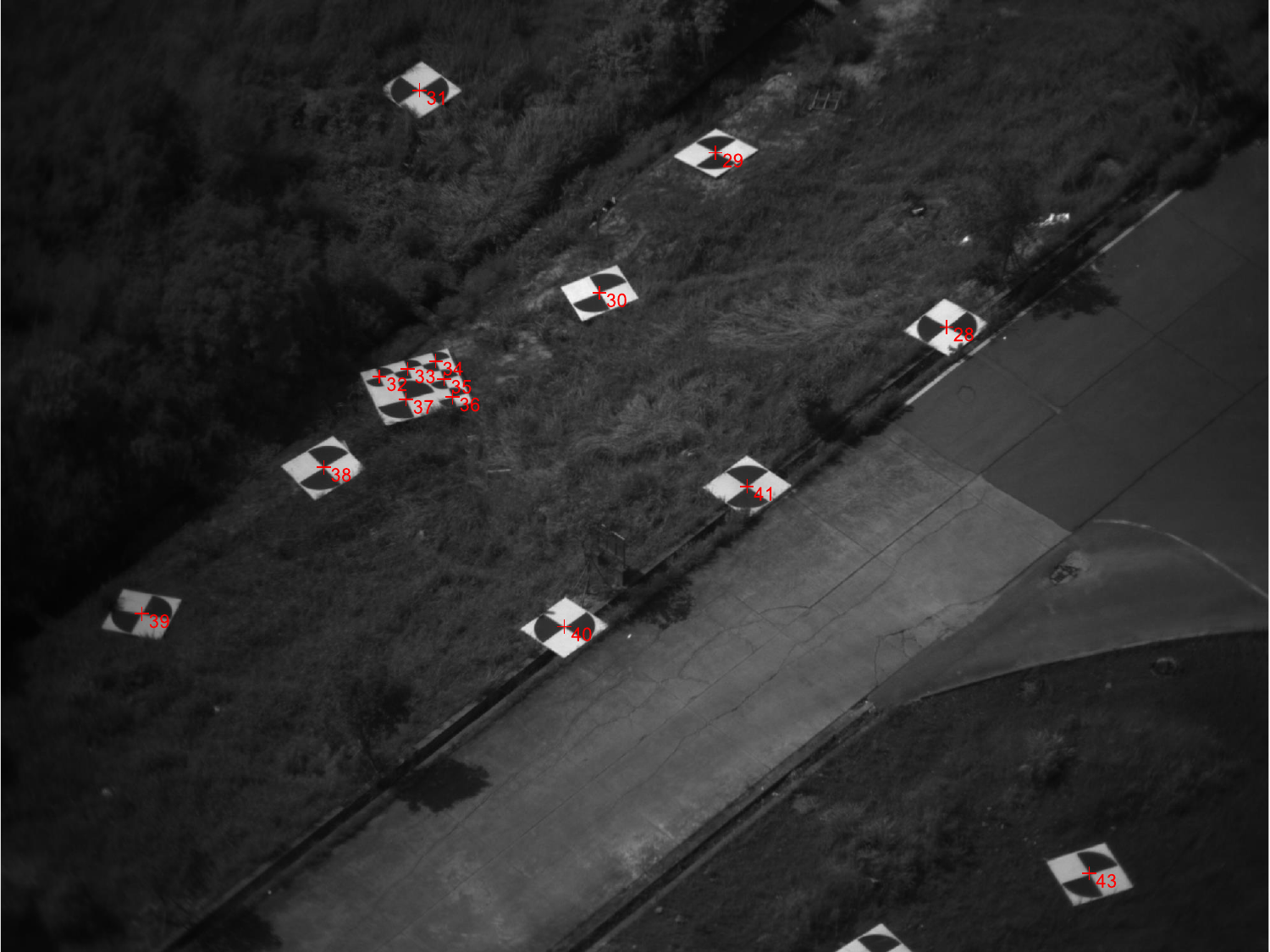} 
	\\
	\vspace{0.1cm}
	\includegraphics[width=0.25\linewidth]{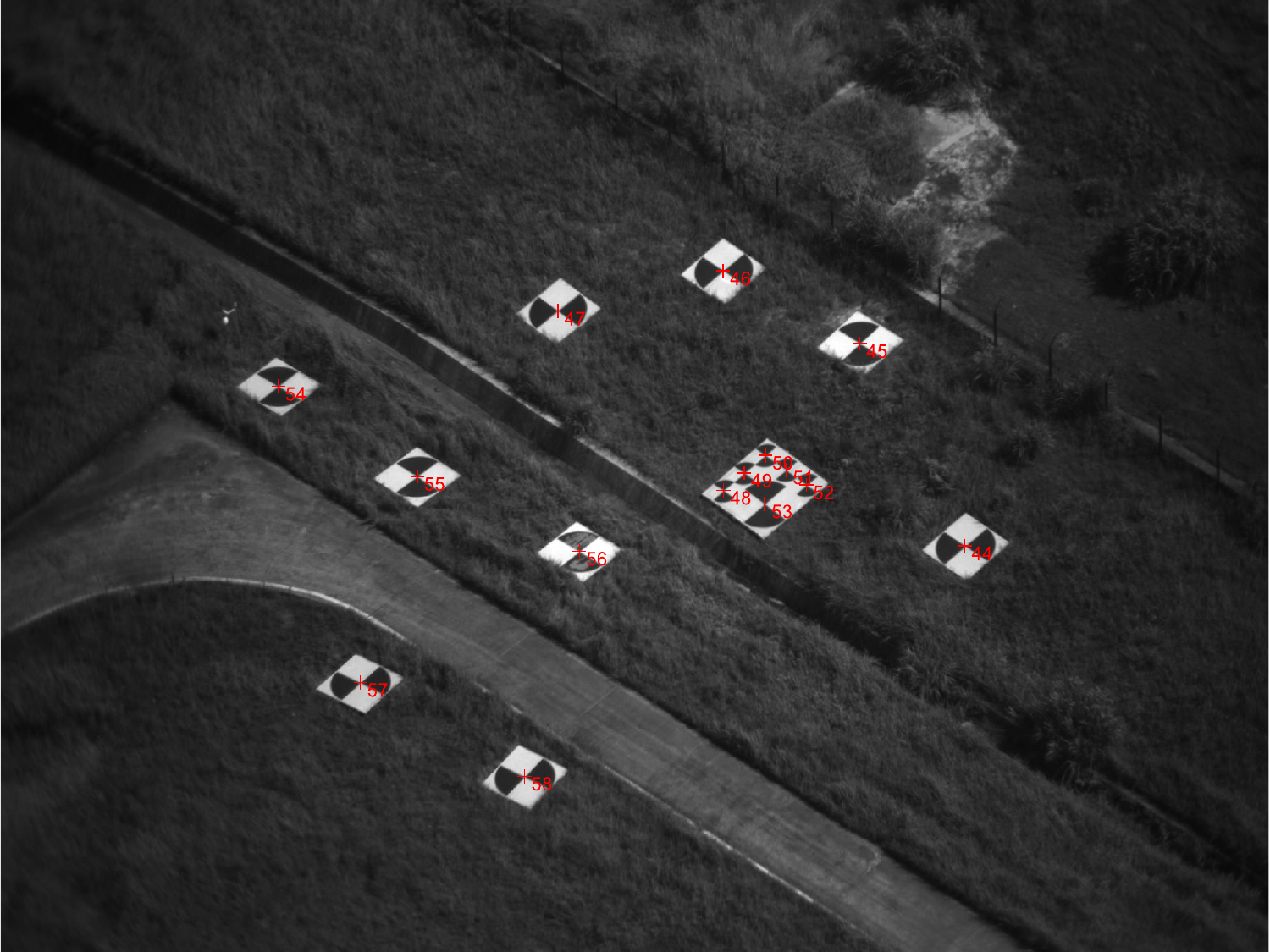} 
	\includegraphics[width=0.25\linewidth]{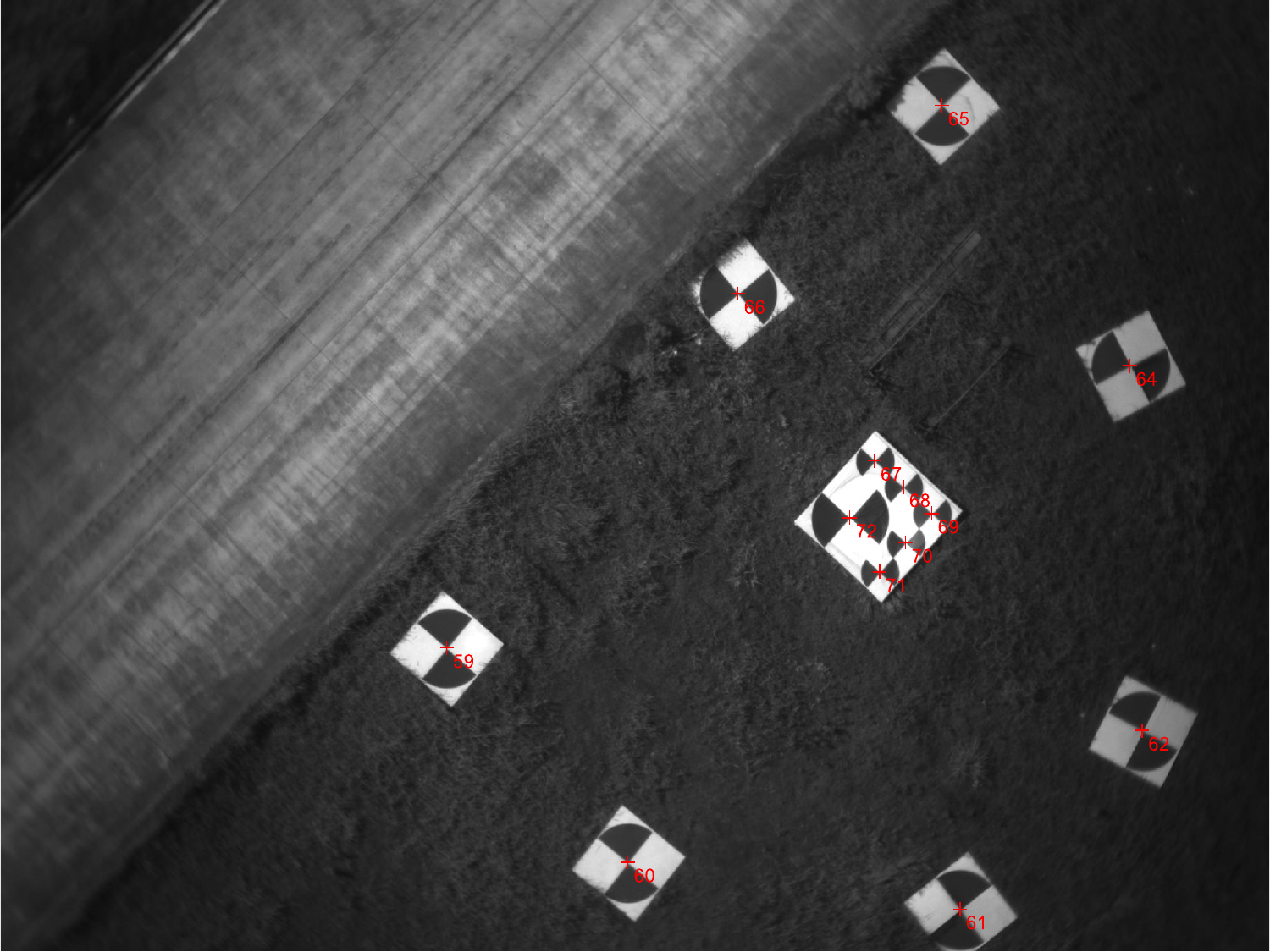} 
	\caption{The sample images obtained from the flight experiment and the results of marker extraction.}
	\label{fig:A}	
\end{figure*}
\begin{figure*}[tbp]
	\centering
	\subfloat[\#0]{
		\includegraphics[width=0.15\linewidth]{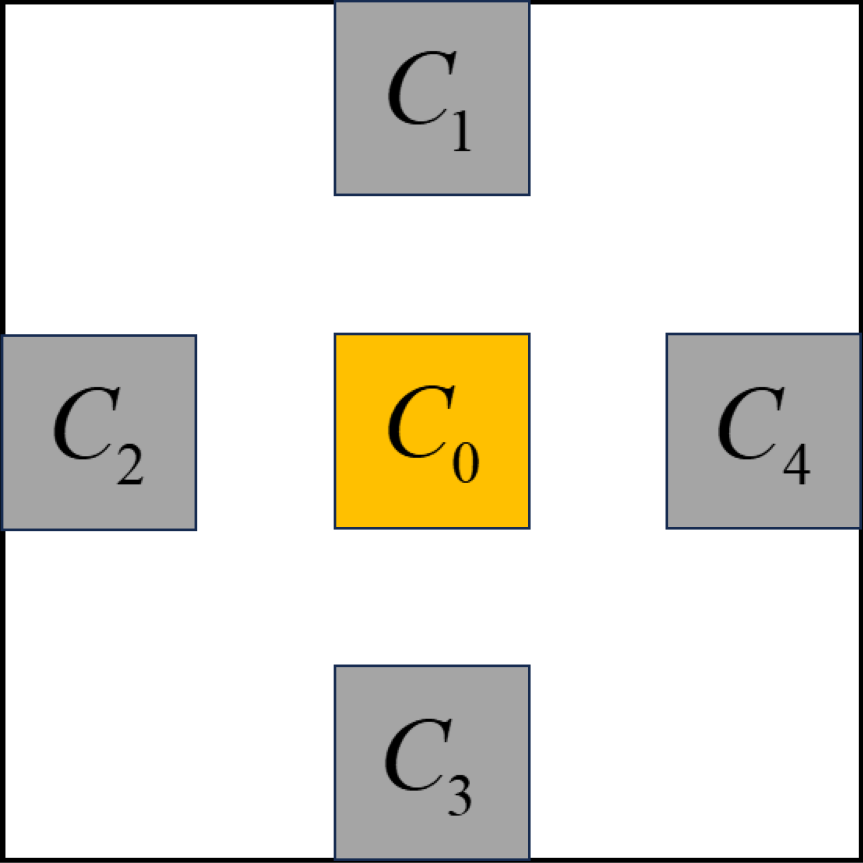} 
		\label{fig:conf0}
		
	}
	\subfloat[\#01]{
		\includegraphics[width=0.15\linewidth]{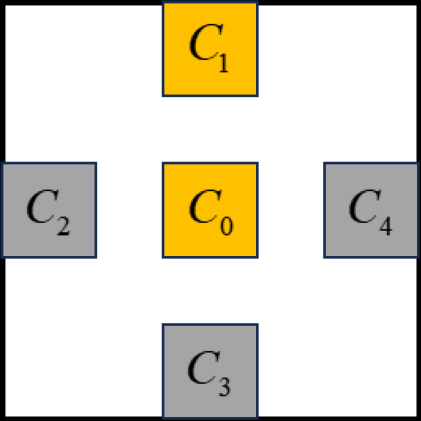}
		\label{fig:conf01}
	}
	\subfloat[\#012]{
		\includegraphics[width=0.15\linewidth]{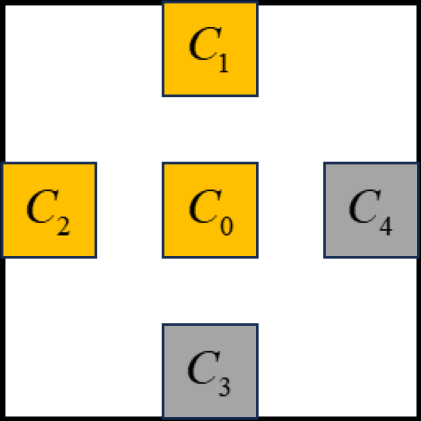} 
		\label{fig:conf012}
	}
	\subfloat[\#1234]{
		\includegraphics[width=0.15\linewidth]{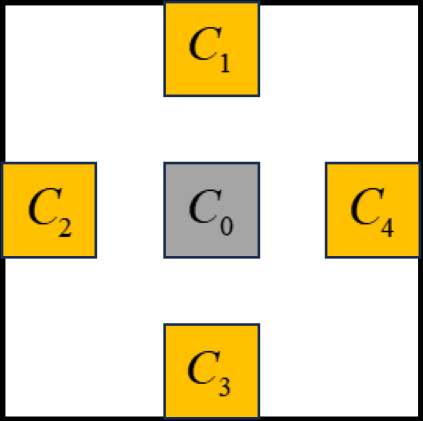} 
		\label{fig:conf0123}
	}
	\subfloat[\#01234]{
		\includegraphics[width=0.15\linewidth]{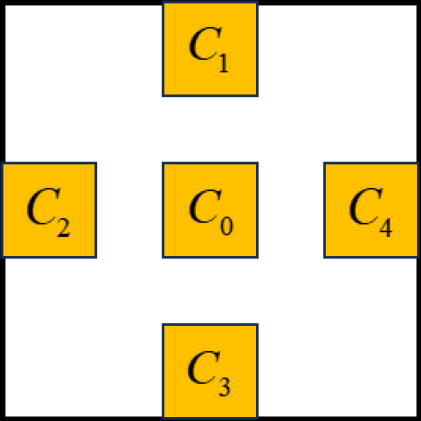} 
		\label{fig:conf01234}
	}
	\caption{Different camera configurations for pose estimation.}
	\label{fig:camera_conf}
\end{figure*}
Once the calibration parameters are obtained, the DMAIS can be applied to the pose estimation task on the flight platform. In this experiment, we evaluate the effectiveness and accuracy of the pose estimation method in real flight scenarios. Our DMAIS is equipped on a four-rotor UAV and observes the ground, as shown in \cref{fig:uav}. Additionally, {a Position and Orientation System (POS) is installed  on the UAV to offer high-precision reference positions.} The spatial position calibration between POS and DMAIS is achieved by calculating the average offset between the estimated position provided by the DMAIS and the position reported by the POS. To conveniently evaluate our pose estimation algorithm, we place several diagonal markers on the ground to provide observation inputs. The position of diagonal markers is measured by a high-precision RTK. The UAV hovered at an altitude of about 350 meters and observed the ground diagonal markers. The computing unit on DMAIS controls five cameras to capture and save images synchronously. {This experiment mainly verifies the accuracy of our pose estimation algorithm. Thus, we process the captured images and evaluate the accuracy of pose estimation offline}. Fig. \ref{fig:A} shows the sample images of markers observed by the DMAIS. We extract the image coordinates of the diagonal marker centers and assign them unique numbers.

We evaluate the accuracy of our pose estimation algorithm in real flight experiment images. The position error $\varepsilon_{pos}$ is calculated as the Euclidean distance between the estimated position and the reference position. Since there is no high-precision orientation ground truth, we use the angular error $\varepsilon_{ang}$ to indirectly evaluate the pose estimation error. $\varepsilon_{ang}$ can be calculated as the product of re-projection error $\varepsilon_{proj}$ and angular resolution. The angular resolution of the camera can be calculated via intrinsic parameters. The units for position error $\varepsilon_{pos}$ are meters ($m$), for angular error $\varepsilon_{ang}$ are arc-minute ($^\prime$) and for re-projection error $\varepsilon_{proj}$ are pixels $(px)$. We conducted five groups of flight experiments at two different locations, each involving multiple pose estimations. The mean and standard deviation of error are used to evaluate the pose estimation results, detailed in \cref{tab:pose_methods}. {The average position error of the DMAIS is about $0.1m$, with a standard deviation of $0.04$, achieving centimeter-level positioning in most scenarios. The angle error of all groups is less than $0.5^\prime$, with a standard deviation of less than $0.1^\prime$, indicating that DMAIS can achieve arc-minute-level orientation estimation.} 

{The eighth column of \cref{tab:pose_methods} demonstrates the runtime of our method. We implemented our absolute pose estimation algorithm in C++ and deployed it on the onboard computer. The computer is configured with an Intel Core i7 with 2.5 GHz. The average running time is less than 0.5 milliseconds, which far exceeds the requirements for real-time performance at the 30 Hz frame rate. Additionally, to better validate the runtime of our algorithm, we also performed runtime simulations with different numbers of points. The results are shown in \cref{tab:time_points}. The runtime of our method increases correspondingly with the number of points. Even with 500 points, our method can still maintain real-time performance at 30 Hz frame rate.} 
\begin{table}[tbp]
	\centering
	\caption{{Pose estimation results in five flight experiments.}}
	\resizebox{\linewidth}{!}{
		\begin{tabular}{cccccccc}
			\toprule
			\multirow{2}{*}{Groups} & \multicolumn{2}{c}{$\varepsilon_{pos}(m)$} & \multicolumn{2}{c}{$\varepsilon_{ang}(^\prime)$} & \multicolumn{2}{c}{$\varepsilon_{proj}(px)$} & \multirow{2}{*}{{Time($ms$)}}\\
			\cline{2-7}
			& mean & std & mean & std & mean & std & \\
			\midrule
			{\ding{172}} & 0.061 & 0.026 & 0.323 & 0.067 & 4.112 & 0.856 & {0.52}\\
			{\ding{173}} & 0.059 & 0.031 & 0.330 & 0.091 & 4.202 & 1.164 & {0.44}\\
			{\ding{174}} & 0.095 & 0.030 & 0.325 & 0.079 & 4.137 & 1.001 & {0.48}\\
			{\ding{175}} & 0.054 & 0.026 & 0.313 & 0.075 & 3.986 & 0.954 & {0.53}\\
			{\ding{176}} & 0.120 & 0.061 & 0.425 & 0.067 & 5.148 & 0.854 & {0.43}\\
			\bottomrule
		\end{tabular}
		\	}
	\normalsize
	\label{tab:pose_methods}
\end{table}

\begin{table}[ht]
	\centering
	\caption{{Average Running Time as the Number of Points Increases.}}
	\begin{tabular}{cccccccc}
		\toprule
		{Points} & 10 & 50 & 100 & 200 & 300 & 400 & 500 \\
		\midrule
		{Times ($ms$)} & 0.34 & 0.44 & 0.55 & 0.75 & 0.96 & 1.14 & 1.36 \\
	    \bottomrule
	\end{tabular}
	\label{tab:time_points}
\end{table}

\begin{table}[tbp]
	\centering
	\caption{Pose estimation results with different camera configurations.}
	\resizebox{\linewidth}{!}{
		\begin{tabular}{ccccccc}
			\toprule
			\multicolumn{2}{c}{Groups}& \#0 & \#01 & \#012 & \#1234 & \#01234 \\
			\midrule
			\multirow{3}{*}{\ding{172}} & $\varepsilon_{pos}(m)$ & 1.413 & 0.545 & 0.064 & 0.071 & \textbf{0.061} \\
			& $\varepsilon_{ang}(\prime)$ & 4.517 & 1.419 & 0.345 & 0.327 & \textbf{0.323} \\
			& $\varepsilon_{proj}(px)$ & 57.572 & 18.093 & 4.398 & 4.170 & \textbf{4.112} \\
			\hline
			\multirow{3}{*}{\ding{173}} & $\varepsilon_{pos}(m)$ & 0.889 & 0.939 & \textbf{0.035} & 0.081 & 0.059 \\
			& $\varepsilon_{ang}(\prime)$ & 2.477 & 2.248 & 0.341 & 0.340 & \textbf{0.330} \\ 
			& $\varepsilon_{proj}(px)$ & 31.576 & 28.648 & 4.344 & 4.334 & \textbf{4.202} \\ 
			\hline
			\multirow{3}{*}{\ding{174}} & $\varepsilon_{pos}(m)$ & 1.062 & 0.397 & 0.104 & 0.101 & \textbf{0.095} \\
			& $\varepsilon_{ang}(\prime)$ & 3.386 & 1.037 & 0.346 & 0.328 & \textbf{0.325} \\ 
			& $\varepsilon_{proj}(px)$ & 43.145 & 13.213 & 4.404 & 4.183 & \textbf{4.137} \\  
			\hline
			\multirow{3}{*}{\ding{175}} & $\varepsilon_{pos}(m)$ & 1.336 & 0.741  & 0.059 & 0.065  & \textbf{0.054} \\
			& $\varepsilon_{ang}(\prime)$ & 5.559 & 1.850 & 0.344 & 0.317 & \textbf{0.313} \\  
			& $\varepsilon_{proj}(px)$ & 70.877 & 23.591 & 4.381 & 4.041 & \textbf{3.986} \\ 
			\hline
			\multirow{3}{*}{\ding{176}} & $\varepsilon_{pos}(m)$ & 1.633 & 0.147 & \textbf{0.103} & 0.125 & 0.120 \\
			& $\varepsilon_{ang}(\prime)$ & 7.202 & \textbf{0.406} & 0.469 & 0.434 & 0.425 \\
			& $\varepsilon_{proj}(px)$ & 91.832 & \textbf{5.173} & 5.985 & 5.536 & 5.418 \\   			
			\bottomrule
		\end{tabular}
	}
	\normalsize
	\label{tab:conf}
\end{table}
In certain flight scenarios where the camera is obscured by clouds or directly unable to work, the five cameras in the DMAIS cannot contribute to the pose estimation at the same time. This experiment evaluates the pose estimation performance of the DMAIS when only a subset of cameras is working properly. Different camera configurations of the DMAIS are illustrated in \cref{fig:camera_conf}. The yellow-marked camera works normally, while the gray-marked camera is inactive. For example, in the configuration \#012, only the images captured by cameras $(C_0, C_1, C_2)$ are involved in the pose estimation. Results are shown in \cref{tab:conf}. {Under extreme conditions, such as when most cameras in the multi-aperture imaging system fail or are blocked, our method can still achieve pose estimation. Experimental results show that the system requires at least one functional camera to complete pose estimation.} Obviously, as the number of cameras increases, the pose estimation error decreases and tends to be stable. When relying on a single camera for pose estimation (\cref{fig:conf0}), the accuracy is relatively low, with a positioning error of approximately $1m$ and an angle error of less than $10^\prime$. When utilizing more than three cameras for pose estimation, higher precision and stability results are achieved. This improvement is primarily due to the larger observation FoV provided by our DMAIS. As the number of available cameras increases, the overall FoV becomes larger, so the pose estimation accuracy is higher. A single camera has a limited FoV of $5.4^\circ \times 4.0^\circ$, while the FoV of DMAIS is about $90^\circ \times 90^\circ$. Experimental results demonstrate that our DMAIS offers more accurate pose estimation compared to a single camera.

\section{Conclusions}
\label{sec:conclusions}
In this paper, we propose a flight platform pose estimation system using a divergent multi-aperture imaging system. {The DMAIS is equivalent to a single camera, which simultaneously achieves both a large field of view and high spatial resolution. } We then introduce a tailored geometric calibration approach for the DMAIS, using the 3D calibration field with a known structure. Real calibration experiments validate the accuracy and validity of the proposed method. Subsequently, we model the DMAIS as a generalized camera and introduce a novel absolute pose estimation algorithm. This method transforms the pose estimation task into a minimization optimization problem {end} offers a polynomial-based solution. To validate our system, we conduct real flight experiments by mounting the DMAIS on a UAV. The results show that our system can achieve centimeter-level positioning and arc-minute-level orientation. 

{Currently, the DMAIS faces limitations in practical applications. The calibration of intrinsic and extrinsic parameters relies on specific calibration environments and hardware conditions, which limits the efficiency and flexibility of system deployment. In the future, we will focus on exploring online self-calibration methods for DMAIS. The system can dynamically utilize natural features in the scene and multi-view geometric constraints to estimate the intrinsic and extrinsic parameters.}

\bibliographystyle{IEEEtran}
\bibliography{LSKbib}
\end{document}